\pdfoutput=1
\documentclass[runningheads]{llncs}

\usepackage{microtype}
\usepackage{amsfonts,amsmath,amssymb}
\usepackage{bbm}
\usepackage{booktabs}
\usepackage{color}
\usepackage{comment}
\usepackage{dsfont}
\usepackage{graphicx}
\usepackage{makecell}
\usepackage{multirow}
\usepackage{tabularx}
\usepackage{wrapfig}
\usepackage{multirow}
\usepackage{xspace}
\usepackage{comment}
\usepackage{enumitem}
\usepackage{subfigure}
\usepackage{floatrow}
\usepackage[pagebackref=true,breaklinks=true,letterpaper=true,colorlinks,bookmarks=false]{hyperref}
\usepackage{cleveref}

\let\originalparagraph\paragraph
\renewcommand{\paragraph}[2][.]{\originalparagraph{#2#1}}

\newcolumntype{L}{>{\raggedright\arraybackslash}X}
\newcolumntype{C}{>{\centering\arraybackslash}X}

\DeclareMathOperator*{\argmin}{argmin}

\makeatletter
\DeclareRobustCommand\onedot{\futurelet\@let@token\@onedot}
\def\@onedot{\ifx\@let@token.\else.\null\fi\xspace}
 
\def\ie{\emph{i.e}\onedot}

\def\etal{\emph{et al}\onedot}
\makeatother

\Crefname{assumption}{\textbf{H}\hspace{-3pt}}{\textbf{H}\hspace{-3pt}}
\crefname{algorithm}{\text{Alg.}}{\text{Alg.}}
\crefname{assumption}{\textbf{H}}{\textbf{H}}
\crefname{equation}{\text{Eq}}{\text{Eq}}
\crefname{definition}{\text{Dfn.}}{\text{Dfn.}}
\crefname{lemma}{\text{Lemma}}{\text{Lemma}}
\crefname{dfn}{\text{Dfn.}}{\text{Dfn.}}
\crefname{thm}{\text{Thm.}}{\text{Thm.}}
\crefname{tab}{\text{Tab.}}{\text{Table}}
\crefname{fig}{\text{Fig.}}{\text{Fig.}}
\crefname{table}{\text{Tab.}}{\text{Table}}
\crefname{figure}{\text{Fig.}}{\text{Fig.}}
\crefname{section}{\text{Sec.}}{\text{Sec.}}

\usepackage[width=122mm,left=12mm,paperwidth=146mm,height=193mm,top=12mm,paperheight=217mm]{geometry}

\usepackage{comment}

\begin{document}
\pagestyle{headings}
\mainmatter
\def\ECCVSubNumber{264}  

\title{Deformation-Aware 3D Model \\
Embedding and Retrieval} 

%
\author{Mikaela Angelina Uy\inst{1}\and
  Jingwei Huang\inst{1}\and
  Minhyuk Sung\inst{2}\and\\
  Tolga Birdal\inst{1}\and
  Leonidas Guibas\inst{1}}
\authorrunning{M. A. Uy et al.}
%
\institute{$^{\text{1 }}$Stanford University \qquad
  $^{\text{2 }}$Adobe Research}
\maketitle

\begin{abstract}
We introduce a new problem of \emph{retrieving} 3D models that are \emph{deformable} to a given query shape and present a novel deep \emph{deformation-aware} embedding to solve this retrieval task.
3D model retrieval is a fundamental operation for recovering a clean and complete 3D model from a noisy and partial 3D scan.
However, given a finite collection of 3D shapes, even the closest model to a query may not be satisfactory. This motivates us to apply 3D model deformation techniques to adapt the retrieved model so as to better fit the query.
Yet, certain restrictions are enforced in most 3D deformation techniques to preserve important features of the original model that prevent a perfect fitting of the deformed model to the query.
This gap between the deformed model and the query induces \emph{asymmetric} relationships among the models, which cannot be handled by typical metric learning techniques. Thus, to retrieve the best models for fitting, we propose a novel deep embedding approach that learns the asymmetric relationships by leveraging location-dependent egocentric distance fields. We also propose two strategies for training the embedding network. We demonstrate that both of these approaches outperform other baselines in our experiments with both synthetic and real data. 
Our project page can be found at \href{https://deformscan2cad.github.io/}{deformscan2cad.github.io}.

\end{abstract}
\section{Introduction}

A fundamental task in 3D perception is the 3D reconstruction, where the shape and appearance of a real object are captured into digital form through a scanning process. The result of 3D scanning is usually imperfect, due to sensor noise, outliers, motion blur, and scanning pattern artifacts. 
Despite the advances in robust techniques for fusing  scans~\cite{Newcombe:2011,Bylow:2013,Whelan:2015,Dai:2017,Dai:2017:BundleFusion,Huang:2017}, the quality of the produced 3D shapes can be far from what is desired.
Recently, we are witnessing growing efforts to replace the observed noisy, cluttered and partial scans with clean geometry, such as artist-created CAD models~\cite{Li:2015,Avetisyan:2019,Avetisyan:2019:Scan2CAD,Dahnert:2019}. In this way, eventually, an entire scene can be virtualized into a set of 3D models that are free of noise, partiality, and scanning artifacts -- while maintaining the semantically valid structure and realistic appearance. One straightforward way to achieve this goal is to replace the sensor data by a known CAD model \textit{retrieved} from an existing repository.

\begin{figure}[t!]
    \centering
    \includegraphics[width=\linewidth]{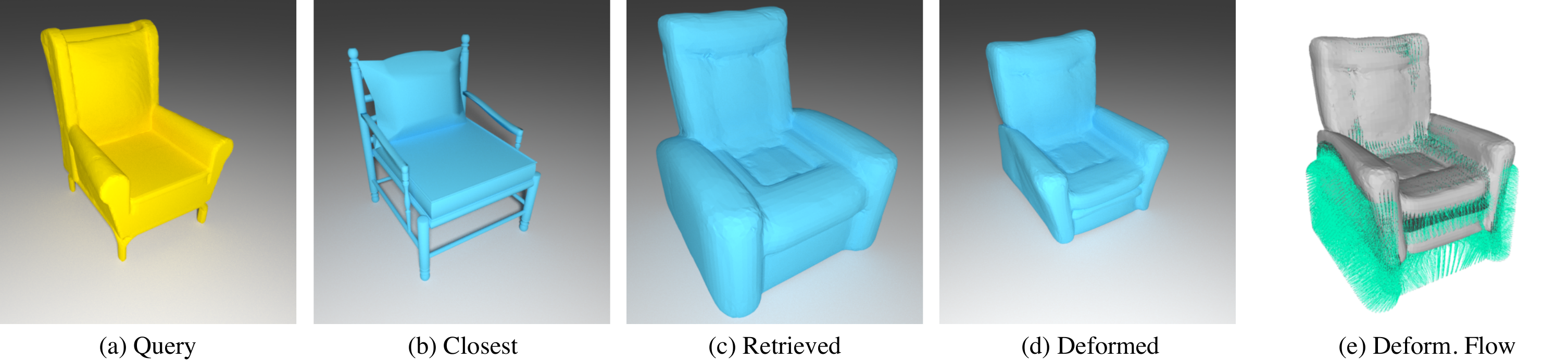}
    \caption{Example of deformation-aware 3D model retrieval.
Given a query (a), the closest 3D model in terms of Chamfer distance has distinct geometric and semantic differences. The model retrieved with our framework (c) 
better fits the query \emph{after} deformation (d). The deformation flow is visualized in (e).}
    \label{fig:best-deformable-example}
\end{figure}

Unfortunately, such retrieval is only viable when there is an almost exact match between the scan and the model.
Given the tremendous variety of real 3D shapes, it is implausible to expect that a CAD model in the repository can exactly match the input or the user's desire -- even with the recent advent of large-scale 3D repositories~\cite{ShapeNet,Warehouse,TurboSquid,GrabCAD}. The closest shape in the database from the query might still have subtle but semantically important geometric or structural differences, leading to an undesirable gap in various settings (e.g., the difference in global structure in~\cref{fig:best-deformable-example} (b)).
To reduce such differences, we propose to retrieve a CAD model (\cref{fig:best-deformable-example} (c)) with similar structure to the query, so that we can apply a deformation operation to fit the query (\cref{fig:best-deformable-example} (d)) better than the closest shape (\cref{fig:best-deformable-example} (b)).
One challenge is to efficiently retrieve such a CAD model especially given that the deformation requires significant time to compute.
In light of this, we propose an efficient \emph{deformation-aware} 3D model retrieval framework that finds a 3D model best matching the input \emph{after} a deformation. Such an approach of joint retrieval and fitting can help more closely reconstruct the target with the same initial pool of 3D models  (\cref{fig:best-deformable-example} (d)) while maintaining retrieval efficiency.


A key issue in this deformation-aware retrieval is in dealing with the \emph{scope} of the deformation of each 3D model. Since the goal of 3D model retrieval is to take advantage of the high fidelity and fine structure of shape representation of man-made models, it is desired to maintain such beneficial properties in the deformation. The long literature of 3D shape deformation has also stemmed from this preservation intent and has investigated diverse ways of constraining or regularizing the deformation; for example, making a smooth function of the deformation with a coarse set of control points~\cite{Sederberg:1986,Kraevoy:2008,Ju:2005,Joshi:2007,Lipman:2008,Weber:2009} or having per-edge or per-face regularization functions preserving local geometric details, given a mesh representation of a model~\cite{Sorkine:2004,Lipman:2004,Lipman:2005,Igarashi:2005,Sorkine:2007}. Such constraints/regularizations aim to ensure production of \emph{plausible} variations without losing the original 3D model's features -- although they simultaneously confine the scope of deformation and prevent it from exactly matching the target. Thus, given a function deforming a \emph{source} model to match the \emph{target} under appropriate regularizations, we consider the notion of the \emph{fitting gap}, defined as the difference between the \emph{deformed} source model and the target.

We introduce a novel deep embedding technique that maps a collection of 3D models into a latent space based on the fitting gap as characterized by a given deformation function. Our embedding technique is agnostic to the exact nature of the deformation function and only requires values of the fitting gap for sampled pairs of the source and target models in training. Due to the \emph{asymmetric} nature of the fitting gap and the lack of a \emph{triangle inequality}, the embedding cannot be accomplished with typical metric learning techniques~\cite{Hadsell:2006,Chechik:2010,Schroff:2015}. Hence, we propose a novel approach, learning a location-dependent \emph{egocentric} anisotropic distance field from the fitting gaps and suggest two network training strategies
: one based on margin loss and the other based on regression loss. In test time, given a query shape, the retrieval can be performed by computing the egocentric distance from all 3D models in the database and finding the one that gives the smallest distance.

In our experiments with ShapeNet~\cite{ShapeNet} dataset, we demonstrate that our framework outperforms all the other baseline methods and also that the second regression-based training strategy provides consistently better performance across different categories of the shapes. We also test our framework with queries of 3D scans and images. In the case of real 3D scans, our outputs show even a smaller average fitting gap when compared with human selected 3D models.

In summary, our contributions are:
\begin{itemize}
    \item defining a new task, that of retrieving a 3D CAD model in a \emph{deformation-aware} fashion;
    \item introducing a novel \emph{asymmetric} distance notion called \emph{fitting gap}, measuring shape difference after deforming one model toward the other;
    \item formulating an \emph{egocentric anisotropic distance field} on a latent embedding space so as to respect the asymmetry of the fitting gap;
    \item proposing two deep network training strategies to learn the said embedding;
    \item demonstrating that our framework outperforms baselines in the experiments with ShapeNet and presenting results for 3D object reconstruction in a real scan-to-CAD application as well as an image-to-CAD scenario.
\end{itemize}

\section{Related Work}
\label{sec:related_work}

\paragraph{\textbf{3D Model Deformation}}
3D model deformation has been a decades-long problem in geometry. Given a shape represented with a mesh, the problem is defined as finding the best positions of vertices in a way that the new shape fits a target while preserving local geometric details.

Previous work has introduced various ways of formulating the regularization conserving the local geometric details, which are mainly classified into three categories.
The first is so-called \emph{free-form}~\cite{Sederberg:1986,Kraevoy:2008} approaches. These methods use the voxel grids of the volume enclosing the surface as control points and define a smooth deformation function interpolating weights from the control points to the mesh vertices.
The second are \emph{cage-based} approaches~\cite{Ju:2005,Joshi:2007,Lipman:2008,Weber:2009}, which take the control points not from voxel grids but from a coarse scaffold mesh surrounding the input.
The last is \emph{vertex-based} approaches~\cite{Sorkine:2004,Lipman:2004,Lipman:2005,Igarashi:2005,Sorkine:2007}. In these methods, the objective function for the optimization is directly defined with the mesh vertex positions, which describe geometric properties that should be preserved, such as mesh Laplacian~\cite{Sorkine:2004,Lipman:2004} or local rigidity~\cite{Lipman:2005,Igarashi:2005,Sorkine:2007}.

Recently, neural networks also have been applied to these three (free-form \cite{Yumer:2016,Hanocka:2018,Jack:2018,Kurenkov:2018}, cage-based~\cite{NeuralCages}, and vertex-based~\cite{3DN,CycleConsistency}) approaches of the 3D shape deformation. The purposes of leveraging neural networks in the deformation vary, including: better handling partiality in the target~\cite{Hanocka:2018}, finding per-point correspondences in an unsupervised way~\cite{CycleConsistency}, enabling taking data in other modalities (e.g., color images or depth scans) as input~\cite{3DN,Jack:2018,Kurenkov:2018}, correlating shape variations with semantic meanings~\cite{Yumer:2016}, and deformation transfer~\cite{NeuralCages}.

In this work, we propose a deformation-aware retrieval technique that can employ any of the deformation methods introduced above as a \emph{given} function for generating plausible deformations. We assume that, based on the regularization of preserving geometric properties, the given deformation function \emph{guarantees} the plausibility of the deformed 3D model while minimizing the fitting distance.
\paragraph{\textbf{Retrieval via Deep Embedding}}
With deep embedding, retrieval problems have been formulated in diverse ways depending on their characteristics.

A significant progress has been made on learning \emph{similarity metrics}, after Chopra~\etal~\cite{Chopra:2005} and Hadsell~\etal~\cite{Hadsell:2006} introduced pioneering work for Siamese network architecture and contrastive loss. Given positive and negative samples of the query, the contrastive loss is defined as pulling and pushing the positive and negative samples in the embedding, respectively. While the contrastive loss is often defined with two separate losses for each of them~\cite{Simo-Serra:2015,Tian:2017}, in the retrieval, considering \emph{relative} distances can engage more flexibility in the embedding function. Thus, later work has more exploited margin losses~\cite{Han:2015,Kumar:2016}, coupling positive and negative samples and pushing the distance between them to be greater than a threshold. Researchers have also verified that the performance can be improved with a better strategy of triplet sampling, such as hard negative mining~\cite{Schroff:2015,Simo-Serra:2015} that takes the farthest positive and the closest negative samples at each time. In~\cref{sec:method_triplet}, we introduce an embedding approach incorporating techniques above, although our problem is fundamentally different from the metric learning due to \emph{asymmetry}. Thus, we focus on dealing with the asymmetry.

Another direction is \emph{graph embedding}, which is more general in terms of handling asymmetric relationships (when considering \emph{directed} graphs). 
The basic goal of the graph embedding is to represent adjacencies among nodes in the graph with similarity in the embedding space. Thus, it can be formulated as regressing the existence or weight of edges~\cite{Ahmed:2013,William:2017}. However, recent work focuses more on learning high-order proximity, with the assumption that \emph{neighbors of neighbors are neighbors}, and leverages ideas of random walk in the embedding~\cite{Perozzi:2014,Grover2016,Dong:2017}. This \emph{transitivity} assumption, however, is \emph{not} guaranteed to hold in our problem. In~\cref{sec:method_regression}, we introduce our second embedding approach following the idea of similarity regression but without exploiting the random walk procedure.

Although metric learning has been previously adapted for 3D point sets~\cite{deng2018ppfnet}, it is shown that non-metric learning is able to generate a more complex, accurate and perceptually meaningful similarity model~\cite{garcia2019learning,tan2006learning}. While similarity search on non-metric spaces is widespread in the classical retrieval systems~\cite{skopal2011nonmetric,chen2008efficient,skopal2008nm,skopal2006fast,morozov2018non}, simultaneous learning and non-metric embedding of deep features is still an open problem. In this paper, we address this gap for the particular problem of deformation-aware 3D shape retrieval.

\section{Deformation-Aware Embedding}
\label{sec:method}
%
%
\setlength{\intextsep}{1pt}
\begin{wrapfigure}{r}{0.5\textwidth}
\centering
    \includegraphics[width=0.9\textwidth]{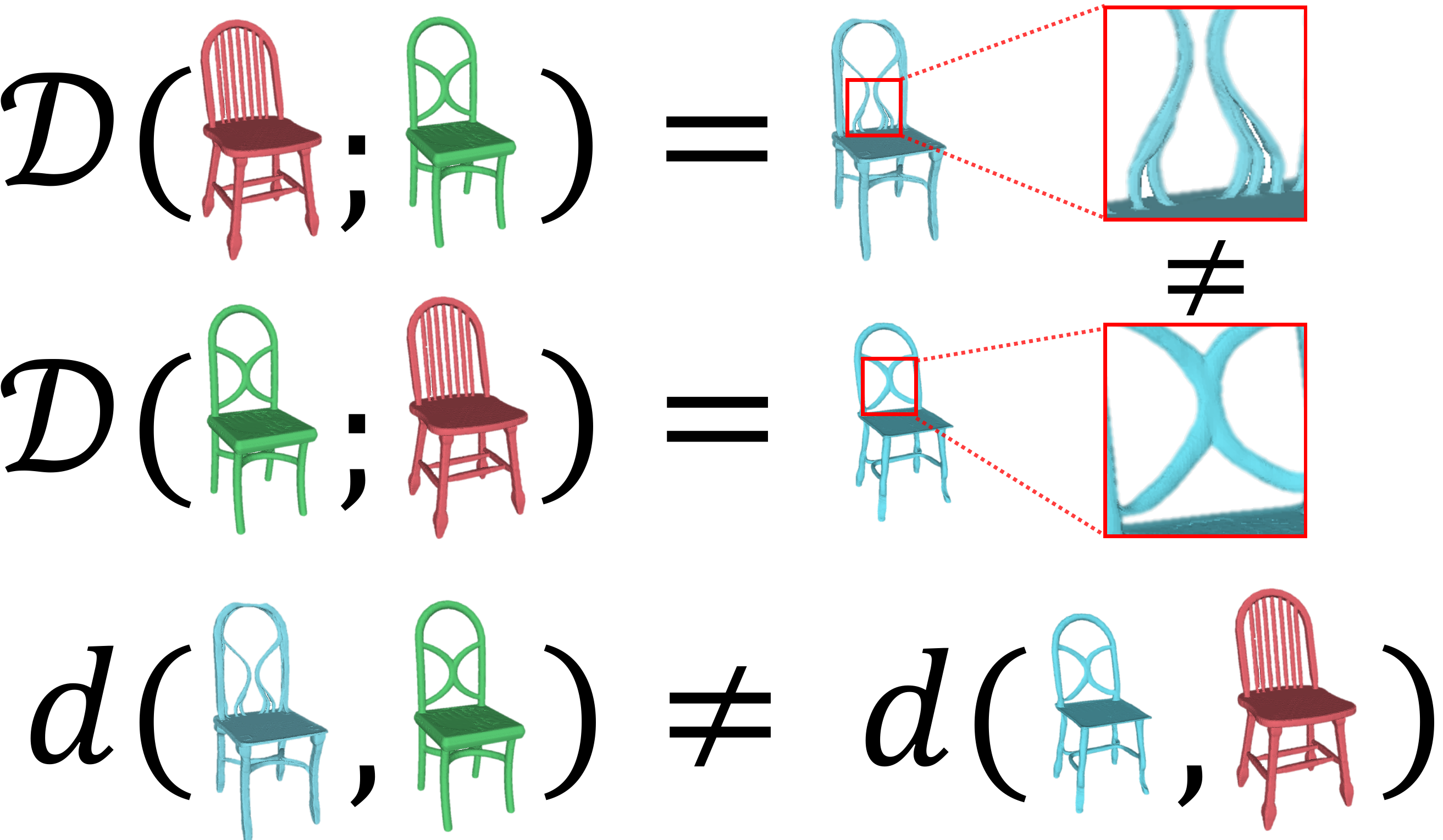}
  \caption{\footnotesize \emph{Fitting gap is asymmetric}. The four bars of the red chair can deform close to the two bars of the green chair,
  achieving a small fitting gap. However, it is harder to deform the green chair into the red chair as we cannot split two bars into four, hence resulting in a larger fitting gap.
  }
  \label{fig:asymmetry}
\end{wrapfigure}
%
We propose an efficient deformation-aware retrieval framework which 
retrieves a 3D model from the database that can best match the query shape through deformation --- in the context of the deformation, we will also use the terms \emph{source} and \emph{target} for the \emph{database} and \emph{query} shapes, respectively.
For the framework, we develop a deep embedding technique that maps a given collection of 3D models $\mathbf{X}$
into a latent space based on a given notion of \textit{distance after deformation}. While in principle any shape can be deformed to any other shape under the same topology, such notion of \emph{fitting gap} emerges from the consideration of constraints or regularizations in the deformation.
A 3D model, which can be easily converted into a mesh, typically has delicate geometric structure that faithfully describes sharp edges and smooth surfaces. Thus, in the deformation of meshes, previous research has paid attention to preserve the fine geometric structure and proposed a variety of techniques regularizing the deformation --  in ways to maintain mesh Laplacian~\cite{Sorkine:2004,Lipman:2004}, local rigidity~\cite{Lipman:2005,Igarashi:2005,Sorkine:2007}, and surface smoothness~\cite{Sederberg:1986,Kraevoy:2008}. Such regularizations, however, obviously may limit the scope of deformation of each 3D model, meaning that a model may not exactly reach the other target shape via deformation. Thus, given a pair of the source and target models $\mathbf{s}, \mathbf{t} \in \mathbf{X}$ and a deformation function $\mathcal{D}: (\mathbf{X} \times \mathbf{X}) \rightarrow \mathbf{X}$ warping the source shape $\mathbf{s}$ to best match the target shape $\mathbf{t}$ under the regularizations, we define the fitting gap $e_\mathcal{D}(\mathbf{s}, \mathbf{t})$ from $\mathbf{s}$ to $\mathbf{t}$ as how much the deformed source shape $\mathcal{D}(\mathbf{s};\mathbf{t})$ \emph{deviates} from the target shape $\mathbf{t}$:
\begin{equation}
    e_\mathcal{D}(\mathbf{s}, \mathbf{t}) = d(\mathcal{D}(\mathbf{s};\mathbf{t}), \mathbf{t}),
    \label{eq:fitting gap}
\end{equation}
where $d: (\mathbf{X} \times \mathbf{X}) \rightarrow [0, \infty)$ is a function measuring the difference between two 3D models.
In other words, the fitting gap is \emph{shape difference after the source deformation} ($0$ means perfect fitting).
Considering the given deformation function $\mathcal{D}$ as a black-box, our goal of the embedding is to build a latent space reflecting the fitting gap characterized by the deformation function so that given a query (target) $\mathbf{t}$,  the 3D model $\mathbf{\hat{s}} \in \mathbf{X} \setminus \{\mathbf{t}\}$ that gives the smallest fitting gap $e_\mathcal{D}(\mathbf{\hat{s}}, \mathbf{t})$ can be retrieved for downstream applications.

Note that such definition of the fitting gap does not guarantee \emph{symmetry} given arbitrary deformation function $\mathcal{D}$: $\exists \, \mathbf{s}, \mathbf{t} \in \mathbf{X} \,\, \text{s.t.} \,\, e_\mathcal{D}(\mathbf{s}, \mathbf{t}) \neq e_\mathcal{D}(\mathbf{t}, \mathbf{s})$; a counterexample can be found as shown in~\cref{fig:asymmetry}. Moreover, any notion of transitivity such as directional triangular inequality ($e_\mathcal{D}(\mathbf{s}, \mathbf{t}) + e_\mathcal{D}(\mathbf{t}, \mathbf{u}) \leq e_\mathcal{D}(\mathbf{s}, \mathbf{u})$)
is not guaranteed. For both reasons, the fitting gap is not a \emph{metric}.
The only properties of metrics that are satisfied with the fitting gap are the following two:
\begin{enumerate}[noitemsep]
    \item (Non-negativity) $e_\mathcal{D}(\mathbf{s}, \mathbf{t}) \geq 0$ for every $\mathbf{s}, \mathbf{t} \in \mathbf{X}$.
    \item (Identity) $e_\mathcal{D}(\mathbf{t}, \mathbf{t}) = 0$ for every $\mathbf{t} \in \mathbf{X}$.~\footnote{This is \emph{not} exactly the same with the property of metrics, \emph{identity of indiscernibles}, meaning the two-way identity ($e_\mathcal{D}(\mathbf{s}, \mathbf{t}) = 0 \Leftrightarrow \mathbf{s} = \mathbf{t}$). We cannot guarantee that $e_\mathcal{D}(\mathbf{s}, \mathbf{t}) = 0 \Rightarrow \mathbf{s} = \mathbf{t}$ from our definition of $e_\mathcal{D}$. Nevertheless, this is not necessary in the retrieval problem.}
\end{enumerate}
Non-negativity holds since $d$ in~\cref{eq:fitting gap} is a distance function. For identity, we assume that the given deformation function $\mathcal{D}$ satisfies $\mathcal{D}(\mathbf{t}, \mathbf{t}) = \mathbf{t}$ (making no change when the source and target are the same), and thus $e_\mathcal{D}(\mathbf{t}, \mathbf{t}) = d(\mathcal{D}(\mathbf{t}), \mathbf{t}) = d(\mathbf{t}, \mathbf{t}) = 0$.
A family of such bivariate functions is often called pseudosemimetrics~\cite{Buldygin:2000} or premetrics~\cite{Aldrovandi:1995}.
Embedding based on such a notion has been underexplored.

Next, we illustrate how we encode the fitting gap among 3D models on a latent embedding space (\cref{sec:method_mahalanobis}) and then propose two strategies of training our embedding network (\cref{sec:method_triplet,sec:method_regression}).

\subsection{Embedding with Egocentric Distances}
\label{sec:method_mahalanobis}

\begin{figure}[t]
    \centering
    \subfigure[]{
    \centering
    \includegraphics[width=0.49\linewidth]{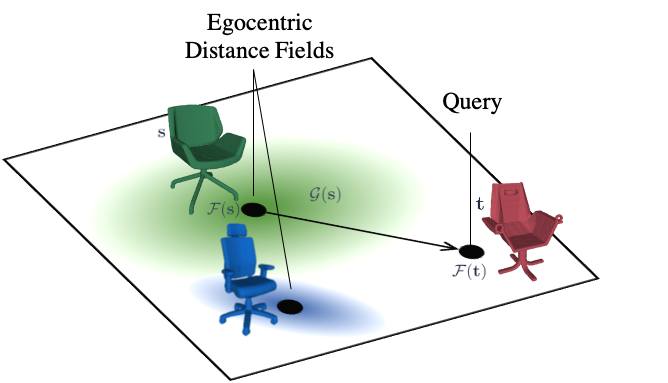}
    \label{fig:embedding}
    }\hfill
     \subfigure[]{
    \centering
    \includegraphics[width=0.45\linewidth]{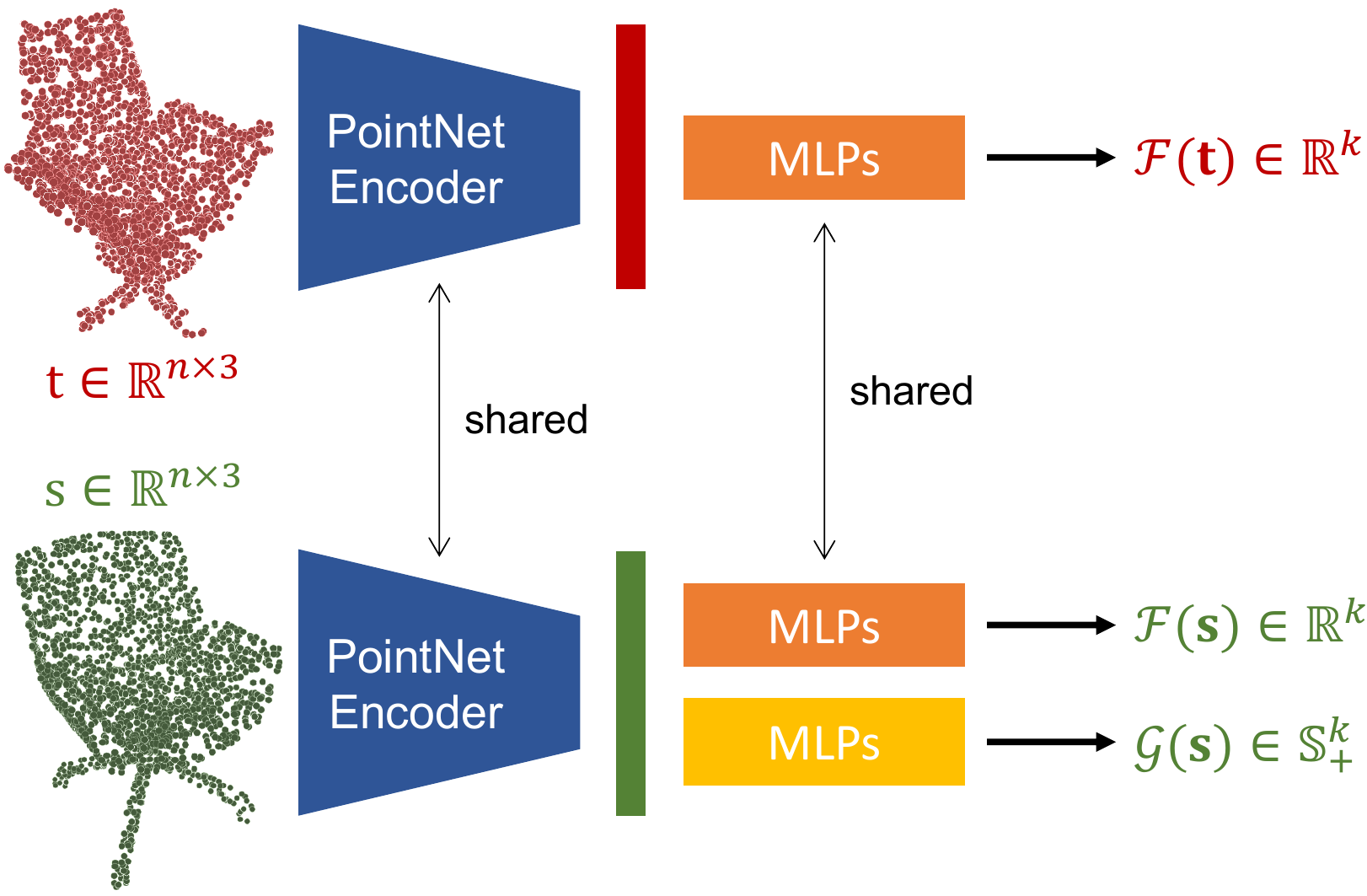}
    \label{fig:network}
   }
    \caption{\textbf{(a)} Visual illustration of our embedding space and egocentric distance. \textbf{(b)} Our Siamese network architecture. The PointNet~\cite{qi2016pointnet} encoder branches to two MLPs: the embedding network $\mathcal{F}$ and the egocentric distance field network $\mathcal{G}$. Both the embedding vector $\mathcal{F}(\mathbf{s})$ and the distance field $\mathcal{G}(\mathbf{s})$ are predicted for the source $\mathbf{s}$, while only the embedding vector $\mathcal{F}(\mathbf{t})$ is predicted for the target $\mathbf{t}$. These are used to calculate for their asymmetric fitting gap.}
\end{figure}

Consider an embedding network $\mathcal{F}: \mathbf{X} \rightarrow \mathbb{R}^{k}$ that maps each 3D model in $\mathbf{X}$ to a point in a $k$-dimensional latent space.
The key in our embedding is to allow the network to properly encode \emph{asymmetric} relationships among 3D models described with the fitting gap while satisfying the properties including non-negativity and identity.
Given this, in addition to mapping a 3D model $\mathbf{s} \in \mathbf{X}$ to a point in the embedding space, we propose another network $\mathcal{G}: \mathbf{X} \rightarrow \mathbb{S}^{k}_{+}$ that predicts an \emph{egocentric} anisotropic distance field for each 3D model, represented with a $k \times k$ positive-semidefinite (PSD) matrix. Analogous to Mahalanobis distance~\cite{Mahalanobis:1936}, we define the egocentric distance function $\delta: \mathbf{X} \times \mathbf{X} \rightarrow [0, \infty)$ given the target and \emph{observer} 3D models $\mathbf{t}, \mathbf{s}\in \mathbf{X}$ as follows:
\begin{align}
    \delta(\mathbf{t}; \mathbf{s}) = \sqrt{\left( \mathcal{F}(\mathbf{t}) - \mathcal{F}(\mathbf{s}) \right)^{T} \mathcal{G}(\mathbf{s}) \left( \mathcal{F}(\mathbf{t}) - \mathcal{F}(\mathbf{s}) \right) }.
    \label{eq:egocentric_distance}
\end{align}
Although it is a common practice to employ Mahalanobis distance in metric learning~\cite{Liu:2012,Bellet:2013,Kulis:2013},
we do \emph{not} learn a metric. Hence, we propose to vary the PSD matrix (the inverse covariance matrix in the Mahalanobis distance) depending on the \emph{observer} shape
so that it can characterize the fitting gap of the observer shape over the latent space. We remark that, in retrieval, each model in the database that can be \emph{retrieved} becomes an \emph{observer} when computing the distance from the query to the model since we deform the retrieved 3D model to fit the query (see~\cref{fig:embedding}).
Also, note that the function $\delta$ satisfies non-negativity (since $\mathcal{G}(\mathbf{s})\succeq 0$) and identity ($\forall \mathbf{s}, \,\, \delta(\mathbf{s}; \mathbf{s}) = 0$).

When considering the goal of retrieval (\cref{sec:method}), our desire is to learn a egocentric distance function $\delta$ that satisfies for every $t$ that
\begin{align}
\argmin_{\mathbf{s} \in \mathbf{X} \setminus \{\mathbf{t}\}} e_\mathcal{D}(\mathbf{s}, \mathbf{t}) = \argmin_{\mathbf{s} \in \mathbf{X} \setminus \{\mathbf{t}\}} \delta(\mathbf{t}; \mathbf{s}).
\end{align}
Since it is practically impossible to compute the deformation function $\mathcal{D}$ for all ordered pairs of 3D models in $\mathbf{X}$ due to the intensive computation time,
we leverage the inductive bias of neural networks generalizing the prediction to unseen data points.
Thus, in network training, we select a fixed size subset of models $\mathbf{X}_{\mathbf{t}} \setminus \{\mathbf{t}\} \subset \mathbf{X}$ for every model $\mathbf{t} \in \mathbf{X}$ and only use the set of source-target pairs $\{ (\mathbf{s}, \mathbf{t}) \,|\, \mathbf{s} \in \mathbf{X}_{\mathbf{t}}, \forall \mathbf{t} \in \mathbf{X} \}$ in the training while precomputing the fitting gap.
In the following subsections, we introduce two training strategies with different loss functions. The difference between these two strategies is also analyzed with experiments in~\cref{sec:exp}.


\subsection{Margin-Loss-Based Approach}
\label{sec:method_triplet}
We first propose our margin-loss-based approach, inspired by previous weakly supervised learning work~\cite{Arandjelovi:2016}.
We leverage on having a notion of positive (deformable) and negative (not deformable) candidates for each query shape. For a query (target) shape $\mathbf{t}$, we define a positive set $\mathbf{P}_\mathbf{t} = \{ \mathbf{s} \in \mathbf{X}_\mathbf{t} \, | \, e_\mathcal{D}(\mathbf{s}, \mathbf{t}) \leq \sigma_{\mathbf{P}} \}$ and a negative set $\mathbf{N}_\mathbf{t} = \{ \mathbf{s} \in \mathbf{X}_\mathbf{t} \, | \, e_\mathcal{D}(\mathbf{s}, \mathbf{t}) > \sigma_{\mathbf{N}} \}$ of the 3D models based on the thresholds $\sigma_{\mathbf{P}}$ and $\sigma_{\mathbf{N}}$ ($\sigma_{\mathbf{P}} < \sigma_{\mathbf{N}}$). In training, we sample triplets $\left(\mathbf{t}, \mathbf{P}^\prime_\mathbf{t}, \mathbf{N}^\prime_\mathbf{t} \right)$ by taking random subsets $\mathbf{P}^\prime_\mathbf{t} \subset \mathbf{P}_\mathbf{t}$ and $\mathbf{N}^\prime_\mathbf{t} \subset \mathbf{N}_\mathbf{t}$ and define the loss as follows:
\begin{equation}
\mathcal{L}_{M}\left(\mathbf{t}, \mathbf{P}^\prime_\mathbf{t}, \mathbf{N}^\prime_\mathbf{t} \right) =
\frac{1}{\mathbf{N}^\prime_\mathbf{t}} \sum_{\mathbf{n} \in \mathbf{N}^\prime_\mathbf{t}} [ \max_{\mathbf{p} \in \mathbf{P}^\prime_\mathbf{t}}
\left( \delta\left(\mathbf{t};\mathbf{p}\right) \right) - \delta\left(\mathbf{t};\mathbf{n}\right) + m]_{+},
\end{equation}
where $[\ldots]_{+}$ denotes the hinge loss~\cite{cortes1995support} and $m$ is a margin parameter.
This is in contrast to the loss of Arandjelovi~\etal~\cite{Arandjelovi:2016} where the best/closest positive is taken to handle false positives.
The intuition for our loss is that the distance from the query to the furthest positive candidate should always be pulled closer than any of the negative candidates.

\subsection{Regression-Based Approach}
\label{sec:method_regression}
We also propose another training strategy that uses a regression loss instead of defining the positive and negative sets.
Since we only need to learn \emph{relative} scales
of the fitting gap
$e_\mathcal{D}(\mathbf{s},\mathbf{t})$ for each query $\mathbf{t}$ in retrieval, inspired by Stochastic Neighbor Embedding (SNE)~\cite{Hinton:2003}, we first convert the fitting gap into a form of \emph{probability} as follows:

\begin{equation}
p(\mathbf{s}; \mathbf{t}) = \frac{\exp\left(-e_\mathcal{D}^2(\mathbf{s},\mathbf{t})/2\sigma_\mathbf{t}^2\right)}{\sum_{\mathbf{s} \in \mathbf{X}^\prime_\mathbf{t}}{\exp\left(-e_\mathcal{D}^2(\mathbf{s},\mathbf{t})/2\sigma_\mathbf{t}^2\right)}}, 
\end{equation}
where $\mathbf{X}^\prime_\mathbf{t} \subset \mathbf{X}_\mathbf{t}$ is a randomly sampled fixed size subset and $\sigma_\mathbf{t}$ is a pre-computed constant for each shape $\mathbf{t}$, which is determined in a way to satisfy the following condition (which is based on Shannon entropy)~\cite{VanDerMaaten:2008}:
\begin{equation}
\log_2 \tau=-\sum_{\mathbf{s} \in \mathbf{X}^\prime_\mathbf{t}} p(\mathbf{s}; \mathbf{t}) \log_2(p(\mathbf{s}; \mathbf{t})),
\end{equation}
where $\tau$ is a perplexity parameter determining the extent of the neighborhood.
Note that we regress the probabilities $p(\mathbf{s}; \mathbf{t})$ since we do not have access to the entire distribution of $p(\cdot; \mathbf{t})$ but only to the models in the subset $\mathbf{X}^\prime_\mathbf{t}$ for each $\mathbf{t}$. This is contrast to SNE which seeks to fully match the given and predicted distributions.
We similarly convert the learned asymmetric distance $\delta(\mathbf{t}; \mathbf{s})$ into a form of probability:
\begin{equation}
\hat{p}(\mathbf{s}; \mathbf{t}) = \frac{\delta^2(\mathbf{t}; \mathbf{s})}{\sum_{\mathbf{s} \in \mathbf{X}^\prime_\mathbf{t}}{\delta^2(\mathbf{t}; \mathbf{s})}}.
\end{equation}
The following $l1$-distance is finally defined as regression loss:
\begin{equation}
\mathcal{L}_{R}(\mathbf{t}, \mathbf{X}^\prime_\mathbf{t}) = \frac{1}{\mathbf{X}^\prime_\mathbf{t}} \sum_{\mathbf{s} \in \mathbf{X}^\prime_\mathbf{t}} | \hat{p}(\mathbf{s}; \mathbf{t}) - p(\mathbf{s}; \mathbf{t}) |.
\end{equation}
\section{Implementation Details}
\label{sec:implementation_details}

In our implementation, we first convert 3D CAD models into meshes to compute deformation.
In order to cope with the multiple connected components of the CAD models in the deformation, we particularly convert the CAD models into \emph{watertight} meshes using the method of Huang~\etal~\cite{huang2018robust}. We further simplify them with a mesh decimation technique~\cite{garland1998simplifying} for efficient computation.
For the deformation function $\mathcal{D}$, we use a simplified version of ARAP deformation~\cite{Sorkine:2007} --- refer to the supplementary for the details. For computing the distance between shapes and feeding the 3D shape information to the network, we also generate point clouds from the meshes by uniformly sampling $2,048$ points. The distance function $d(\mathbf{x},\mathbf{y})$ (see~\cref{eq:fitting gap}) measuring the shape difference between two 3D models $\mathbf{x}, \mathbf{y} \in \mathbf{X}$ is defined as average two-way Chamfer distance (CD) between the point sets resampled on the meshes $\mathbf{x}, \mathbf{y}$ following previous work~\cite{Fan2016APS,Achlioptas:2018,3DN,NeuralCages}. We remark that our embedding framework does \emph{not} require any specific type of the deformation function, and in the embedding, the given deformation function is only used to precompute the fitting gap between two shapes in the sampled pairs. See \cref{sec:related_work} for more options of the deformation function.

\paragraph{\textbf{Network Architecture for $\mathcal{F}$ and $\mathcal{G}$ and training details}}
~\cref{fig:network} illustrates our network design.
We build a Siamese architecture taking a pair of source $\mathtt{s}$ and target $\mathtt{t}$ point clouds with PointNet~\cite{qi2016pointnet} encoder (the earlier part until the \textit{maxpool} layer) as our shared encoder.
The outputs after the \textit{maxpool} then pass through two separate branches of \textit{MLP};
one is $\mathcal{F}$ that predicts the location in the $k$-dimensional latent embedding space, and the other is $\mathcal{G}$ that predicts the egocentric distance field (the PSD matrix, see ~\cref{sec:method_mahalanobis}).
In $\mathcal{G}$, we predict a \emph{positive diagonal} matrix as our PSD matrix using a sigmoid activation and adding $\epsilon =1e^{-6}$.
We use $k=256$ for most of our experiments but also demonstrate the effect of varying the dimension of the latent space in the supplementary material.

In training, we further randomly downsample the point clouds with $2,048$ points to $1,024$ points for memory efficiency (but the entire $2,048$ points are used in baseline methods).
We set the minibatch size as 8 for the query $\mathbf{t}$, and $|\mathbf{P'_t}|=2$ and $|\mathbf{N'_t}|=13$ for the margin-loss-based and $|\mathbf{X'_t}|=15$ for the regression-based approaches. We use Adam optimizer with a learning rate of 0.001 and train for 350 epochs for all cases. 

Remark that the resolution of the 3D model in retrieval is not affected by the resolution of input point clouds fed into our network. A 3D model in any resolution can be retrieved in their original format. 
\section {Results}
\label{sec:exp}
We present our experimental evaluation to demonstrate the advantage of our embedding framework for deformation-aware 3D model retrieval. We also showcase applications of our approach in two real scenarios: Scan-to-CAD (\cref{sec:exp_scan2cad}) and Image-to-CAD\footnote{Due to space restrictions we present results of Image-to-CAD in our supplementary material.}.



\paragraph{\textbf{Baselines}} We compare the proposed margin-loss-based (\emph{Ours-Margin},~\cref{sec:method_triplet}) and regression-based (\emph{Ours-Reg},~\cref{sec:method_regression}) approaches with three retrieval baselines (we also compare with more baselines in the supplementary):
\begin{enumerate}
    \item Ranked by Chamfer Distance (\emph{Ranked CD}): This retrieves the closest 3D models by Chamfer Distance (CD), which is our distance function $d$ in~\cref{sec:implementation_details}.
    \item Autoencoder (\emph{AE}): This learns an embedding space by training a point cloud autoencoder as defined in~\cite{Achlioptas:2018}.
    The dimension of the latent space is $1024$, which is larger than that of our space.
    \item Chamfer Distance Triplet (\emph{CD-Margin}):
    This baseline is the same with \emph{Ours-Margin} (\cref{sec:method_triplet}) except for that the distance for the hinge loss is defined as the Euclidean distance over the latent space instead of our egocentric asymmetric distance. The positive and negative candidates are sampled by taking $20$ closest models ordered by CD and random $50$ models, respectively.
\end{enumerate}

The Chamfer Distance Triplet (\emph{CD-Margin}) is trained in the same way with our margin-loss-based approach (\emph{Ours-Margin}) described in~\cref{sec:method_triplet}; the minibatches are generated with 8 queries and 2 positive and 13 negative random candidates for each of them. We use a margin value $m=0.5$ for \emph{CD-Margin} and also normalize the latent codes to have a unit $l2$-norm as done in FaceNet~\cite{Schroff:2015}.

Note that neither of the three baselines above leverage the information about \emph{deformability}, meaning how a 3D model can be deformed to fit the query.

\setlength{\tabcolsep}{4pt}
\begin{table}[t!]
\centering
\caption{Quantitative results of retrievals. See \cref{sec:exp} for baselines and evaluation metrics. The numbers multiplied by $1e^{-2}$ are reported. Bold is the smallest, and underscore is the second smallest. Our retrieval results give smaller \emph{after}-deformation distances $e^m_\mathcal{D}(\mathbf{s},\mathbf{t})$ while the \emph{before}-deformation distances $d^m(\mathbf{s},\mathbf{t})$ are large.}
\label{table:quantitative}
{\scriptsize
\begin{tabularx}{\textwidth}{c|>{\centering}m{2cm}|CC|CC|CC|CC|CC}
\toprule
 \multicolumn{2}{c|}{\multirow{2}{*}{Method}} & \multicolumn{2}{c|}{Table} & \multicolumn{2}{c|}{Chair} & \multicolumn{2}{c|}{Sofa} & \multicolumn{2}{c|}{Car} & \multicolumn{2}{c}{Plane} \\
 \multicolumn{2}{c|}{} & {\tiny Top-1} & {\tiny Top-3} & {\tiny Top-1} & {\tiny Top-3} & {\tiny Top-1} & {\tiny Top-3} & {\tiny Top-1} & {\tiny Top-3} & {\tiny Top-1} & {\tiny Top-3} \\
\midrule
\multirow{5}{*}{\rotatebox{90}{\makecell{Mean\\$d^m(\mathbf{s},\mathbf{t})$}}}
& Ranked CD  & \textbf{4.467} & \textbf{3.287} & \textbf{4.412} & \textbf{3.333} & \textbf{3.916} & \textbf{2.985} & \textbf{2.346} & \textbf{1.860} & \textbf{2.530} & \textbf{1.540}\\
& AE  & \underline{4.867} & 3.491 & \underline{4.710} & \underline{3.473} & 4.223 & \underline{3.178} & 2.579 & 1.942 & 3.045 & 1.789\\
& CD-Margin &  4.875 & \underline{3.449} & 4.750 & 3.518 & \underline{3.087} & 4.151 & \underline{2.525} & \underline{1.905}& \underline{2.801} & \underline{1.655}\\
& \textbf{Ours-Margin} &  6.227 & 4.026 &  5.664 & 3.889 & 4.825 & 3.400  &  2.962 & 2.142 & 3.442 & 1.885\\
& \textbf{Ours-Reg} &  5.955 & 3.979 & 5.751 & 3.981  & 5.091 & 3.628 & 3.119 & 2.263 & 3.436 & 1.976\\
\midrule
\multirow{5}{*}{\rotatebox{90}{\makecell{Mean\\$e^m_\mathcal{D}(\mathbf{s},\mathbf{t})$}}}
& Ranked CD & \underline{2.095}  & 1.284 & 1.937 & 1.186 & 1.450 & 0.886 & \underline{1.138} & \underline{0.716} & \textbf{1.199} & \underline{0.569} \\
& AE  &  2.180 & 1.292  & 1.991 & 1.196 & 1.521 & 0.887  & 1.214 & 0.753 & 1.392 & 0.634 \\
& CD-Margin & 2.362 & 1.373  & 2.134 & 1.242  &  1.587 & 0.909 &  1.249 & 0.773 & 1.315 & 0.620 \\
& \textbf{Ours-Margin} & 2.127 & \underline{1.251} & \underline{1.915} & \underline{1.144} & \underline{1.420} & \underline{0.835} & 1.226 & 0.747 & \underline{1.300} & 0.586 \\
& \textbf{Ours-Reg} &  \textbf{1.969} & \textbf{1.129} & \textbf{1.752} & \textbf{1.054} & \textbf{1.338} & \textbf{0.788} & \textbf{1.112} & \textbf{0.681} & \textbf{1.199} & \textbf{0.529} \\
\bottomrule
\end{tabularx}
}
\end{table}
\setlength{\tabcolsep}{1.4pt}

\paragraph{\textbf{Evaluation Metrics}}
To avoid sampling bias of the point clouds, in the evaluations, we measure the distance between two shapes as a two-way \emph{point-to-mesh} Chamfer distance; we also use a denser point cloud including $50k$ uniformly sampled points in this case. We denote this new distance function as $d^m: (\mathbf{X} \times \mathbf{X}) \rightarrow [0, \infty)$ and the accompanying fitting gap function as $e_\mathcal{D}^m(\mathbf{s}, \mathbf{t}) = d^m(\mathcal{D}(\mathbf{s};\mathbf{t}), \mathbf{t})$.
Also, for simplicity, we will use the notations $d^m(\mathbf{s},\mathbf{t})$ and $e^m_\mathcal{D}(\mathbf{s},\mathbf{t})$ as the source-target distances \emph{before} and \emph{after} the source deformation in the rest of the paper.
We report the mean of these numbers for the best among top-$N$ retrieved models; $N=1, 3$ are reported.
For a partial scan input, we use a \emph{one-way} point-to-mesh distance; see \cref{sec:exp_scan2cad} for the details.

\begin{figure}[t]
    \centering
    \includegraphics[width=\linewidth]{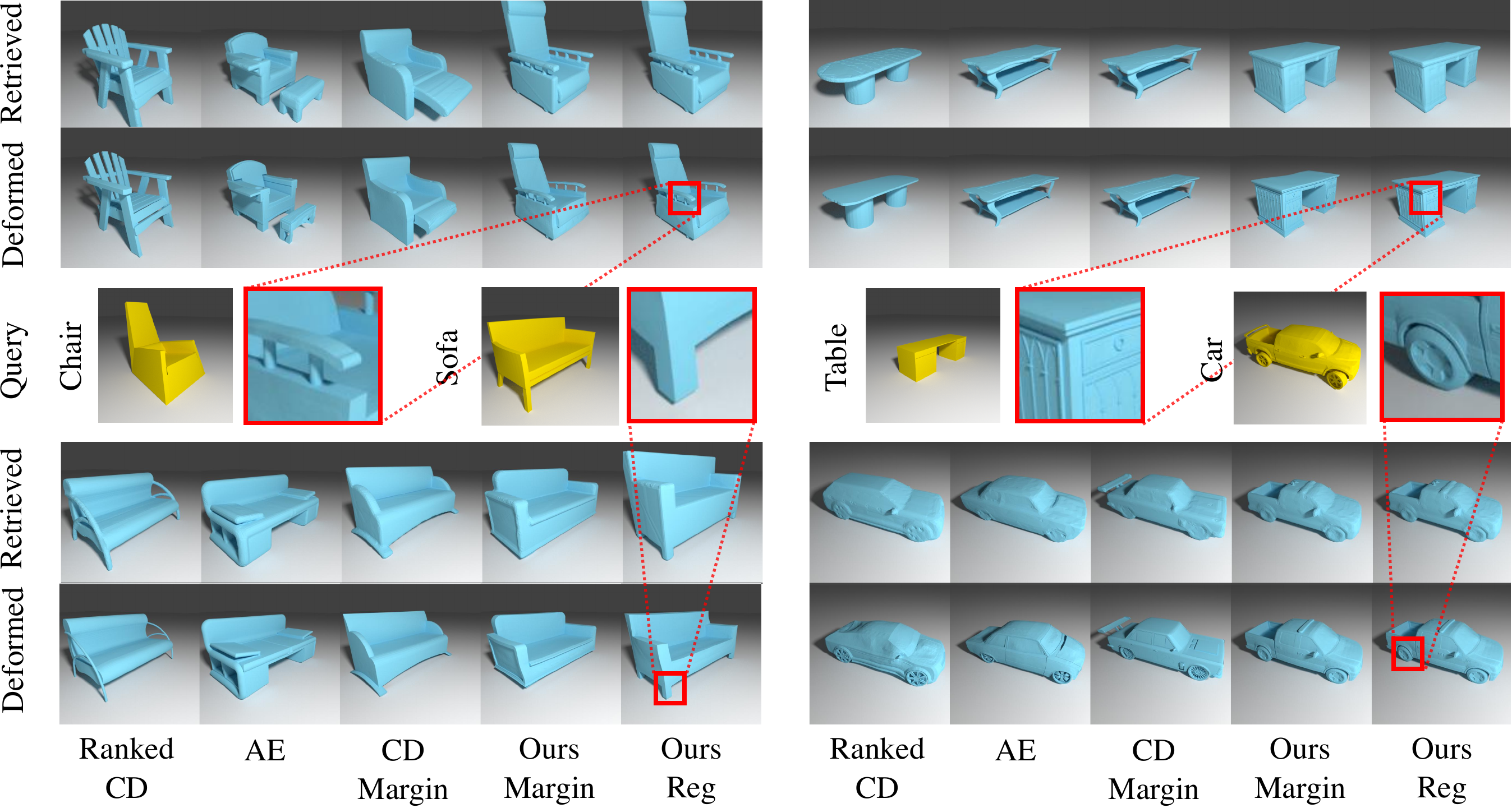}
    \caption{Visualization of retrieval followed by deformation on ShapeNet. Our network is able to retrieve models that better fit after deformation despite having large geometric distances initially. Notice the big back part of the retrieved chair and the thick seat of the retrieved sofa, attributes that are not identical to the query. Yet, these parts properly fit the target after deformation. Our network is also able to retrieve a sofa with legs and a car with a trunk that are present in the desired targets. Moreover, our deformation-aware \emph{retrieval \& deformation} approach also allows us to preserve fine-details of the source model post-deformation as shown in the zoomed in regions. See supplementary for more results.
    }
    \label{fig:shapenet-vis}
\end{figure}

\subsection{Experiments on ShapeNet~\cite{ShapeNet}}
\label{sec:exp_shapenet}
We experiment with four classes in ShapeNet~\cite{ShapeNet} dataset: \emph{Table}, \emph{Chair}, \emph{Sofa} and \emph{Car}.
We train/evaluate the networks per class with the training/test splits of Yang~\etal~\cite{Yang:2019}.
In the evaluations, we take all models in the test set as queries and retrieve 3D models from the same test set but except for the query.
For our training data, we precompute the fitting gap $e_\mathcal{D}$ (\cref{sec:method}). To obtain source-target pairs, we sample 100 source models for every target $\mathbf{t}\in\mathbf{X}$, i.e. $|\mathbf{X_t}|=100$, which consist of the 50 closest models by the distance $d$ in~\cref{sec:method_mahalanobis} (not including $\mathbf{t}$ itself) and another 50 random models. We use $\sigma_P=3.5e^{-4}, 3e^{-4}, 2e^{-4},$ and $1.2e^{-4}$ and $\sigma_N=7.5e^{-4}, 6e^{-4}, 4e^{-4},$ and $2e^{-4}$ for the table, chair, sofa, and car classes, respectively, and $m=10$ for our margin-loss-based approach. We use $\tau=5$ for all classes in our regression-based approach.

We report mean $d^m(\mathbf{s},\mathbf{t})$ and mean $e^m_\mathcal{D}(\mathbf{s},\mathbf{t})$
for all methods in \cref{table:quantitative}.
When observing the results of our two methods, margin-loss-based (\emph{Ours-Margin}) and regression-based (\emph{Ours-Reg}) approaches, the distance \emph{before} deformation ($d^m(\mathbf{s},\mathbf{t})$) is farther than the baselines, but the distance \emph{after} deformation ($e^m_\mathcal{D}(\mathbf{s},\mathbf{t})$) is smaller. Such a result is shown consistently in all classes, particularly for \emph{Ours-Reg} results. This indicates that our methods can discover 3D models that can better align with the query shape through the given deformation operation.

Also, \emph{Ours-Reg} achieves better results than \emph{Ours-Margin} consistently for all classes.
The advantage of \emph{Ours-Reg} is that it can discriminate among all models in $\mathbf{X_t}$, while \emph{Ours-Margin} can only consider \emph{inter-}relationships across the positive set $\mathbf{P_t}$ and the negative set $\mathbf{N_t}$ but not \emph{intra-}relationships in each set.
Hence, \emph{Ours-Reg} can achieve a better understanding of the overall distribution of the data.
\emph{Ours-Margin} also has a trade-off for the thresholds of $\sigma_{\mathbf{P}}$ and $\sigma_{\mathbf{N}}$; too tight thresholds may result in overfitting, and too loose thresholds can make the two sets less distinguishable. We empirically found good thresholds for each class, but finding the optimum thresholds is a very time-consuming task requiring an extensive binary search.

In~\cref{fig:shapenet-vis}, we visualize some examples of retrieved models and their deformations given query models. The models retrieved by our methods have large distance before deformation but better fit after deformation compared with the results of other methods.
For example, the chair and sofa retrieved by our methods as shown in~\cref{fig:shapenet-vis} have bigger back parts than the queries, but they become smaller properly after operating the deformation.
Our network is able to be agnostic to the details that are easy to recover via deformations such as chair body size, and table leg thickness. It rather retrieves based on the overall shape of the model that can be deformed to fit the desired target. On the other hand, the small geometric details can be inherited from the retrieved model and be preserved during deformation. It is also noticeable that our retrieval is more structurally faithful as we observe the presence of legs in the retrieved sofa or the trunk of the car that are essential for valid deformation.



\setlength{\tabcolsep}{0pt}
\begin{table}[t!]
\centering
\label{table:ranking}
\caption{Ranking evaluations with 150 models per query. The models are randomly selected and sorted by $e^m_\mathcal{D}(\mathbf{s},\mathbf{t})$ (the query is not included). All results are for the top-$1$ retrieval results of each method. The numbers multiplied by $1e^{-2}$ are reported. Bold is the smallest, and underscore is the second smallest.
}
\scriptsize
\begin{tabularx}{\textwidth}{l|CCC|CCC|CCC|CCC|CCC}
\toprule
\multirow{3}{*}{Method} & \multicolumn{3}{c|}{Table} & \multicolumn{3}{c|}{Chair} & \multicolumn{3}{c|}{Sofa} & \multicolumn{3}{c|}{Car} & \multicolumn{3}{c}{Plane} \\
 & {\tiny Mean} & {\tiny Mean} & {\tiny Mean}
 & {\tiny Mean} & {\tiny Mean} & {\tiny Mean}
 & {\tiny Mean} & {\tiny Mean} & {\tiny Mean}
 & {\tiny Mean} & {\tiny Mean} & {\tiny Mean} 
 & {\tiny Mean} & {\tiny Mean} & {\tiny Mean} \\
 & {\tiny $d^m$} & {\tiny $e^m_\mathcal{D}$} & {\tiny Rank} 
 & {\tiny $d^m$} & {\tiny $e^m_\mathcal{D}$} & {\tiny Rank} 
 & {\tiny $d^m$} & {\tiny $e^m_\mathcal{D}$} & {\tiny Rank} 
 & {\tiny $d^m$} & {\tiny $e^m_\mathcal{D}$} & {\tiny Rank} 
 & {\tiny $d^m$} & {\tiny $e^m_\mathcal{D}$} & {\tiny Rank} \\
\midrule
Ranked-CD & \textbf{6.24}  & 3.20 & 12.53 & \textbf{5.65}  & 2.61  & 11.37 & \textbf{4.73} & 1.87  & 14.07 & \textbf{2.75} & \underline{1.31}  & \underline{12.0} & \textbf{1.83} & \textbf{1.26} & \textbf{5.53} \\
AE        & 6.95  & 3.11  & 11.69 & 6.08  & 2.61  & 10.21 & 5.19  & 1.91  & 14.43 & 3.09 & 1.39  & 14.55  & 2.60 & 1.68 & 17.91 \\
CD-Margin & \underline{6.77}  & 3.19  & 12.55 & \underline{6.02} & 2.72  & 13.24 & \underline{5.07} & 1.93  & 15.76 & \underline{3.02} & 1.48  & 18.94  & \underline{2.36} & 1.56 & 12.27 \\
\textbf{Ours-Margin} & 8.89  & \underline{2.88} & \underline{8.86} & 7.15 & \underline{2.37} & \underline{8.15} &  5.83 & \underline{1.67} & \underline{9.09} &  3.61 & 1.34 & 12.95  & 2.65 & 1.48 & 10.67 \\
\textbf{Ours-Reg} & 8.59  & \textbf{2.71} & \textbf{7.05} & 7.39  & \textbf{2.24} & \textbf{6.32} &  6.23 & \textbf{1.62} & \textbf{7.91} & 3.80  & \textbf{1.24} & \textbf{7.80}  & 2.64 & \underline{1.42} & \underline{8.96} \\
\bottomrule
\end{tabularx}
\end{table}%

We also report the \emph{rank} of retrieved models when we sort the test models based on $e^m_\mathcal{D}(\mathbf{s},\mathbf{t})$. Since it is computationally extremely expensive to compute the deformation for all pairs in our dataset, for each 3D model, we randomly sample 150 other models and use them for the ranking; $e^m_\mathcal{D}(\mathbf{s},\mathbf{t})$ and $d^m(\mathbf{s},\mathbf{t})$ are precomputed for them.
The results in~\cref{table:ranking} show the mean rank of the top-1 retrieval results out of the selected 150 models.
\emph{Our-Reg} and \emph{Our-Margin} achieve the best and the second best mean ranks compared with the baseline methods in all classes except for cars; most of the car models are structurally similar to each other. \emph{Our-Reg} still provides the best mean rank for cars.
\subsection{Scan-to-CAD}
\label{sec:exp_scan2cad}

\begin{figure}[t!]
    \centering
    \includegraphics[width=\linewidth]{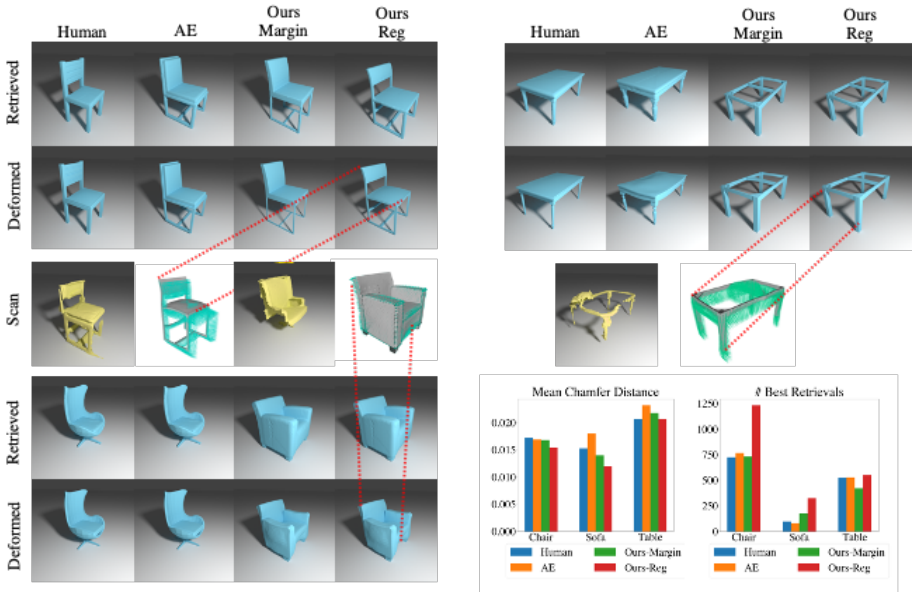}
    \caption{({\textbf{Top row and bottom left}}) Qualitative results of Scan-to-CAD. ({\textbf{Bottom right}}) Quantitative Results: We compare different retrieval methods on the Scan2CAD~\cite{Avetisyan:2019} dataset. The left chart shows the mean fitting errors and the right chart shows the number of best candidates retrieved by different methods. \emph{Ours-Reg} achieves the minimum overall fitting errors and the maximum number of best retrievals among all categories compared with other methods. See supplementary for more results.
    }
    \label{fig:scan2cad-statistics}
\end{figure}

We also evaluate our method for the real scan-to-CAD conversion problem~\cite{Avetisyan:2019}. We use our models trained on ShapeNet~\cite{ShapeNet} and directly evaluate our performance on the Scan2CAD~\cite{Avetisyan:2019} dataset. Scan2CAD provides partial 3D scans of indoor scenes, which are segmented to each object instance. We normalize and align the object scans to the ShapeNet canonical space using the 9DoF alignment provided in~\cite{Avetisyan:2019}.  We use our embedding for retrieval and then apply the deformation function (\cref{sec:implementation_details}) to the retrieved CAD to fit the scan.
Similar to previous works~\cite{3DN,NeuralCages}, we reflect the scan about the vertical symmetric plane before fitting. Our evaluation is performed on three common categories: chairs, sofas, and tables. We compare different methods including human annotations defined by the dataset~\cite{Avetisyan:2019}, the autoencoder, our margin-loss-based retrieval, and our regression-based retrieval. We exclude 149 extremely partial scans (2\% among all scans) which cover less than 10\% of the regions (after reflection) of human-selected shapes, as these shapes are too incomplete to be recognizable.


Similar to the ShapeNet evaluation, we also measure \emph{point-to-mesh} Chamfer distance that is uniformly and densely sampled with 50$k$ points. However, we evaluated on one-way CAD to scan fitting, since our goal is to fit the complete CAD models to all observed regions in the partial scans. The fitting error of \emph{Ours-Reg} is the least in all categories as shown on the left chart on the bottom right of~\cref{fig:scan2cad-statistics}. We show a clear advantage for the chairs and the sofas. For tables, we visualized and found that Scan2CAD tables are quite similar to each other, so many potential candidates can fit the scan relatively well with our chosen deformation function (\cref{sec:implementation_details}).

We also report the number of the best-retrieved models among the evaluated methods  as shown in the right chart on the bottom-right corner of~\cref{fig:scan2cad-statistics}. The number of best candidates with \emph{Ours-Reg} is significantly higher than the baselines and even compared with human-selected models on chairs and sofas.

\cref{fig:scan2cad-statistics} (top row and bottom left) shows the qualitative comparison for different methods in three categories. We see that \emph{Ours-Reg} retrieves models that are more similar to that of the target scan. For example, ours is the only method that retrieves the model with the same connectivity of parts compared to the target as shown by the legs of the chair and the top-less table scan. For the sofa example, we retrieve a model with a large distance but which is similar in shape to the observed regions of the scan, hence the deformed model fits the scan better than other methods. By deforming the CAD model, we additionally preserve important CAD features including sharp edges and corners.

\section {Conclusion}
We proposed a \emph{deformation-aware} 3D model retrieval framework that enables finding a 3D model best matching a query after the deformation. Due to the feature-preserving regularizations in most deformation techniques, 3D models generally cannot exactly match the query through deformation but induce a \emph{fitting gap}. This gap, describing the \emph{deformability} one to the other, is \emph{asymmetric} by definition and thus not a metric. Hence, we introduced a novel embedding technique that can encode such relationships with \emph{egocentric} distance fields given any arbitrary deformation function and proposed two strategies for network training. We demonstrated that our approach outperforms other baselines in the experiments with ShapeNet and also presented results in scan-to-CAD and image-to-CAD applications.
We plan to further investigate the relationships among 3D models defined by the deformation in the future.

\bigbreak
\noindent\textbf{Acknowledgements} This work is supported by a Google AR/VR University Research Award, a Vannevar Bush Faculty Fellowship, a grant from the Stanford SAIL Toyota Research Center, and gifts from the Adobe Corporation and the Dassault Foundation.
\clearpage
\newpage
%
%
{
\bibliographystyle{splncs04}
\bibliography{egbib}

\begin{thebibliography}{10}
\providecommand{\url}[1]{\texttt{#1}}
\providecommand{\urlprefix}{URL }
\providecommand{\doi}[1]{https://doi.org/#1}

\bibitem{Achlioptas:2018}
Achlioptas, P., Diamanti, O., Mitliagkas, I., Guibas, L.J.: Learning
  representations and generative models for 3d point clouds. In: ICML (2018)

\bibitem{ceres-solver}
Agarwal, S., Mierle, K., Others: Ceres solver. \url{http://ceres-solver.org}

\bibitem{Ahmed:2013}
Ahmed, A., Shervashidze, N., Narayanamurthy, S., Josifovski, V., Smola, A.J.:
  Distributed large-scale natural graph factorization. In: WWW (2013)

\bibitem{Aldrovandi:1995}
Aldrovandi, R., Pereira, J.: An Introduction to Geometrical Physics. World
  Scientific (1995)

\bibitem{Arandjelovi:2016}
Arandjelovi\'c, R., Gronat, P., Torii, A., Pajdla, T., Sivic, J.: {NetVLAD}:
  {CNN} architecture for weakly supervised place recognition. In: CVPR (2016)

\bibitem{Avetisyan:2019:Scan2CAD}
Avetisyan, A., Dahnert, M., Dai, A., Savva, M., Chang, A.X., Nie{\ss}ner, M.:
  {Scan2CAD}: Learning cad model alignment in {RGB-D} scans. In: CVPR (2019)

\bibitem{Avetisyan:2019}
Avetisyan, A., Dai, A., Nie{\ss}ner, M.: End-to-end cad model retrieval and
  9dof alignment in 3d scans. In: ICCV (2019)

\bibitem{Bellet:2013}
Bellet, A., Habrard, A., Sebban, M.: A survey on metric learning for feature
  vectors and structured data (2013)

\bibitem{Buldygin:2000}
Buldygin, V., Kozachenko, V., Kozachenko, I., Kozacenko, J., Kozachenko, Y.,
  Koza{\v{c}}enko, {\^U}., Kozachenko, V., Kozachenko, I., Za{\^a}c, V.: Metric
  Characterization of Random Variables and Random Processes. American
  Mathematical Society (2000)

\bibitem{Bylow:2013}
Bylow, E., Sturm, J., Kerl, C., Kahl, F., Cremers, D.: Real-time camera
  tracking and 3d reconstruction using signed distance functions. In: RSS
  (2013)

\bibitem{ShapeNet}
Chang, A.X., Funkhouser, T.A., Guibas, L.J., Hanrahan, P., Huang, Q., Li, Z.,
  Savarese, S., Savva, M., Song, S., Su, H., Xiao, J., Yi, L., Yu, F.:
  Shapenet: An information-rich 3d model repository (2015)

\bibitem{Chechik:2010}
Chechik, G., Sharma, V., Shalit, U., Bengio, S.: Large scale online learning of
  image similarity through ranking. Journal of Machine Learning Research
  (2010)

\bibitem{chen2008efficient}
Chen, L., Lian, X.: Efficient similarity search in nonmetric spaces with local
  constant embedding. IEEE Transactions on Knowledge and Data Engineering
  \textbf{20}(3) (2008)

\bibitem{Chopra:2005}
Chopra, S., Hadsell, R., LeCun, Y.: Learning a similarity metric
  discriminatively, with application to face verification. In: CVPR (2005)

\bibitem{cortes1995support}
Cortes, C., Vapnik, V.: Support-vector networks. Machine learning  (1995)

\bibitem{Dahnert:2019}
Dahnert, M., Dai, A., Guibas, L., Nie{\ss}ner, M.: Joint embedding of 3d scan
  and cad objects. In: ICCV (2019)

\bibitem{Dai:2017:BundleFusion}
Dai, A., Nie{\ss}ner, M., Zollh{\"o}fer, M., Izadi, S., Theobalt, C.:
  {BundleFusion}: Real-time globally consistent 3d reconstruction using
  on-the-fly surface reintegration. In: ACM SIGGRAPH (2017)

\bibitem{Dai:2017}
Dai, A., Ruizhongtai~Qi, C., Nie{\ss}ner, M.: Shape completion using
  3d-encoder-predictor cnns and shape synthesis. In: CVPR (2017)

\bibitem{deng2018ppfnet}
Deng, H., Birdal, T., Ilic, S.: {PPFNet}: Global context aware local features
  for robust 3d point matching. In: Proceedings of the IEEE Conference on
  Computer Vision and Pattern Recognition. pp. 195--205 (2018)

\bibitem{Dong:2017}
Dong, Y., Chawla, N.V., Swami, A.: {metapath2vec}: Scalable representation
  learning for heterogeneous networks. In: KDD (2017)

\bibitem{Fan2016APS}
Fan, H., Su, H., Guibas, L.J.: A point set generation network for 3d object
  reconstruction from a single image. In: CVPR (2016)

\bibitem{Kumar:2016}
G, V.K.B., Carneiro, G., Reid, I.: Learning local image descriptors with deep
  siamese and triplet convolutional networks by minimizing global loss
  functions. In: CVPR (2016)

\bibitem{garcia2019learning}
Garcia, N., Vogiatzis, G.: Learning non-metric visual similarity for image
  retrieval. Image and Vision Computing  \textbf{82} (2019)

\bibitem{garland1998simplifying}
Garland, M., Heckbert, P.S.: Simplifying surfaces with color and texture using
  quadric error metrics. In: Visualization (1998)

\bibitem{CycleConsistency}
Groueix, T., Fisher, M., Kim, V.G., Russell, B.C., Aubry, M.: Deep
  self-supervised cycle-consistent deformation for few-shot shape segmentation.
  In: Eurographics Symposium on Geometry Processing (2019)

\bibitem{Grover2016}
Grover, A., Leskovec, J.: node2vec: Scalable feature learning for networks. In:
  KDD (2016)

\bibitem{Hadsell:2006}
{Hadsell}, R., {Chopra}, S., {LeCun}, Y.: Dimensionality reduction by learning
  an invariant mapping. In: CVPR (2006)

\bibitem{Han:2015}
Han, X., Leung, T., Jia, Y., Sukthankar, R., Berg, A.C.: {MatchNet}: Unifying
  feature and metric learning for patch-based matching. In: CVPR (2015)

\bibitem{Hanocka:2018}
Hanocka, R., Fish, N., Wang, Z., Giryes, R., Fleishman, S., Cohen-Or, D.:
  {ALIGNet}: Partial-{Shape} agnostic alignment via unsupervised learning. ACM
  Transactions on Graphics  (2018)

\bibitem{Hinton:2003}
Hinton, G.E., Roweis, S.T.: Stochastic neighbor embedding. In: NIPS (2003)

\bibitem{Huang:2017}
Huang, J., Dai, A., Guibas, L.J., Nie{\ss}ner, M.: {3Dlite}: towards commodity
  3d scanning for content creation. In: ACM SIGGRAPH Asia (2017)

\bibitem{huang2018robust}
Huang, J., Su, H., Guibas, L.: Robust watertight manifold surface generation
  method for shapenet models (2018)

\bibitem{Igarashi:2005}
Igarashi, T., Moscovich, T., Hughes, J.F.: As-rigid-as-possible shape
  manipulation. In: ACM SIGGRAPH (2005)

\bibitem{Jack:2018}
Jack, D., Pontes, J.K., Sridharan, S., Fookes, C., Shirazi, S., Maire, F.,
  Eriksson, A.: Learning free-{Form} deformations for {3D} object
  reconstruction. In: ICCV (2018)

\bibitem{Joshi:2007}
Joshi, P., Meyer, M., DeRose, T., Green, B., Sanocki, T.: Harmonic coordinates
  for character articulation. In: ACM SIGGRAPH (2007)

\bibitem{Ju:2005}
Ju, T., Schaefer, S., Warren, J.: Mean value coordinates for closed triangular
  meshes. In: ACM SIGGRAPH (2005)

\bibitem{Kraevoy:2008}
Kraevoy, V., Sheffer, A., Shamir, A., Cohen-Or, D.: Non-homogeneous resizing of
  complex models. In: ACM SIGGRAPH Asia (2006)

\bibitem{Kulis:2013}
Kulis, B., et~al.: Metric learning: A survey. Foundations and
  Trends{\textregistered} in Machine Learning  (2013)

\bibitem{Kurenkov:2018}
Kurenkov, A., Ji, J., Garg, A., Mehta, V., Gwak, J., Choy, C.B., Savarese, S.:
  {DeformNet}: Free-{Form} deformation network for 3d shape reconstruction from
  a single image. In: WACV (2018)

\bibitem{Li:2015}
Li, Y., Dai, A., Guibas, L., Nie{\ss}ner, M.: Database-assisted object
  retrieval for real-time 3d reconstruction. In: Eurographics (2015)

\bibitem{Lipman:2004}
{Lipman}, Y., {Sorkine}, O., {Cohen-Or}, D., {Levin}, D., {Rossi}, C.,
  {Seidel}, H.P.: Differential coordinates for interactive mesh editing. In:
  Shape Modeling Applications (2004)

\bibitem{Lipman:2008}
Lipman, Y., Levin, D., Cohen-Or, D.: Green coordinates. In: ACM SIGGRAPH (2008)

\bibitem{Lipman:2005}
Lipman, Y., Sorkine, O., Levin, D., Cohen-Or, D.: Linear rotation-invariant
  coordinates for meshes. In: ACM SIGGRAPH (2005)

\bibitem{Liu:2012}
Liu, E.Y., Guo, Z., Zhang, X., Jojic, V., Wang, W.: Metric learning from
  relative comparisons by minimizing squared residual. In: ICDM (2012)

\bibitem{VanDerMaaten:2008}
van~der Maaten, L., Hinton, G.: Visualizing data using {t-SNE}. Journal of
  Machine Learning Research  (2008)

\bibitem{Mahalanobis:1936}
Mahalanobis, P.C.: On the generalized distance in statistics. In: Proceedings
  of the National Institute of Science. National Institute of Science of India
  (1936)

\bibitem{morozov2018non}
Morozov, S., Babenko, A.: Non-metric similarity graphs for maximum inner
  product search. In: Advances in Neural Information Processing Systems (2018)

\bibitem{Newcombe:2011}
Newcombe, R.A., Izadi, S., Hilliges, O., Molyneaux, D., Kim, D., Davison, A.J.,
  Kohi, P., Shotton, J., Hodges, S., Fitzgibbon, A.: {KinectFusion}: Real-time
  dense surface mapping and tracking. In: ISMAR (2011)

\bibitem{Perozzi:2014}
Perozzi, B., Al-Rfou, R., Skiena, S.: Deepwalk: Online learning of social
  representations. In: KDD (2013)

\bibitem{qi2016pointnet}
Qi, C.R., Su, H., Mo, K., Guibas, L.J.: {PointNet}: Deep learning on point sets
  for 3d classification and segmentation. In: CVPR (2017)

\bibitem{Schroff:2015}
Schroff, F., Kalenichenko, D., Philbin, J.: {FaceNet}: A unified embedding for
  face recognition and clustering. In: CVPR (2015)

\bibitem{Sederberg:1986}
Sederberg, T.W., Parry, S.R.: Free-form deformation of solid geometric models.
  In: ACM SIGGRAPH (1986)

\bibitem{Simo-Serra:2015}
Simo-Serra, E., Trulls, E., Ferraz, L., Kokkinos, I., Fua, P., Moreno-Noguer,
  F.: Discriminative learning of deep convolutional feature point descriptors.
  In: ICCV (2015)

\bibitem{skopal2006fast}
Skopal, T.: On fast non-metric similarity search by metric access methods. In:
  International Conference on Extending Database Technology. Springer (2006)

\bibitem{skopal2011nonmetric}
Skopal, T., Bustos, B.: On nonmetric similarity search problems in complex
  domains. ACM Computing Surveys (CSUR)  \textbf{43}(4),  1--50 (2011)

\bibitem{skopal2008nm}
Skopal, T., Loko{\v{c}}, J.: Nm-tree: Flexible approximate similarity search in
  metric and non-metric spaces. In: International Conference on Database and
  Expert Systems Applications. pp. 312--325. Springer (2008)

\bibitem{Sorkine:2007}
Sorkine, O., Alexa, M.: As-rigid-as-possible surface modeling. In: Eurographics
  Symposium on Geometry Processing (2007)

\bibitem{Sorkine:2004}
Sorkine, O., Cohen-Or, D., Lipman, Y., Alexa, M., R\"{o}ssl, C., Seidel, H.P.:
  Laplacian surface editing. In: Eurographics Symposium on Geometry Processing
  (2004)

\bibitem{GrabCAD}
Stratasys: Grab{CAD} community, \url{https://grabcad.com/library}

\bibitem{tan2006learning}
Tan, X., Chen, S., Li, J., Zhou, Z.H.: Learning non-metric partial similarity
  based on maximal margin criterion. In: 2006 IEEE Computer Society Conference
  on Computer Vision and Pattern Recognition (CVPR'06). vol.~1. IEEE (2006)

\bibitem{Tian:2017}
Tian, Y., Fan, B., Wu, F.: {L2-Net}: Deep learning of discriminative patch
  descriptor in euclidean space. In: CVPR (2017)

\bibitem{Warehouse}
Trimble: {3D} warehouse, \url{https://3dwarehouse.sketchup.com/}

\bibitem{TurboSquid}
TurboSquid: Turbo{Squid}, \url{https://www.turbosquid.com/}

\bibitem{3DN}
Wang, W., Ceylan, D., Mech, R., Neumann, U.: {3DN}: 3d deformation network. In:
  CVPR (2019)

\bibitem{Weber:2009}
Weber, O., Ben‐Chen, M., Gotsman, C.: Complex barycentric coordinates with
  applications to planar shape deformation. In: Eurographics (2009)

\bibitem{wen2019pixel2mesh++}
Wen, C., Zhang, Y., Li, Z., Fu, Y.: Pixel2mesh++: Multi-view 3d mesh generation
  via deformation. In: ICCV (2019)

\bibitem{Whelan:2015}
Whelan, T., Leutenegger, S., Salas-Moreno, R.F., Glocker, B., Davison, A.J.:
  {ElasticFusion}: Dense slam without a pose graph. Robotics: Science and
  Systems (2011)

\bibitem{William:2017}
William L.~Hamilton, Rex~Ying, J.L.: Representation learning on graphs: Methods
  and applications. IEEE Data Engineering Bulletin  (2017)

\bibitem{Yang:2019}
Yang, G., Huang, X., Hao, Z., Liu, M.Y., Belongie, S., Hariharan, B.:
  Pointflow: 3d point cloud generation with continuous normalizing flows. In:
  ICCV (2019)

\bibitem{NeuralCages}
Yifan, W., Aigerman, N., Kim, V., Chaudhuri, S., Sorkine-Hornung, O.: Neural
  cages for detail-preserving 3d deformations (2019)

\bibitem{Yumer:2016}
Yumer, E., Mitra, N.J.: Learning semantic deformation flows with 3d
  convolutional networks. In: ECCV (2016)

\end{thebibliography}
}
\clearpage

\renewcommand{\thesection}{A}
\setcounter{table}{0}
\renewcommand{\thetable}{A\arabic{table}}
\setcounter{figure}{0}
\renewcommand{\thefigure}{A\arabic{figure}}
\setcounter{equation}{0}
\renewcommand{\theequation}{A\arabic{equation}}

\newif\ifpaper
\papertrue

\section*{Appendix}
\ifpaper
  \newcommand\refpaper[1]{\unskip}
\else
  \makeatletter
  \newcommand{\manuallabel}[2]{\def\@currentlabel{#2}\label{#1}}
  \makeatother
  \manuallabel{sec:method_mahalanobis}{3.1}
  \manuallabel{sec:method_triplet}{3.2}
  \manuallabel{sec:method_regression}{3.3}
  \manuallabel{sec:implementation_details}{4}
  \manuallabel{sec:exp}{5}
  \manuallabel{sec:exp_shapenet}{5.1}
  \manuallabel{sec:exp_scan2cad}{5.2}
  \manuallabel{sec:exp_image2cad}{5.3}
  \manuallabel{fig:scan2cad-statistics}{6}
  \manuallabel{table:quantitative}{1}
  \manuallabel{table:ranking}{2}
  \manuallabel{eq:fitting gap}{1}
  \manuallabel{eq:egocentric_distance}{2}
  \newcommand{\refpaper}[1]{in the paper}
\fi

\subsection{Details of Deformation Function $\mathcal{D}$}
We opt to use a simple deformation function for $\mathcal{D}$ in our experiments, which is designed to preserve local rigidity similarly with ARAP~\cite{Sorkine:2007} but much simpler yet effective in practice.
Specifically, given a source mesh
$\mathbf{s}=(\mathcal{V}\in\{\mathbb{R}^{3}\}_{1 \cdots N},\mathcal{E}\in\mathcal{V}^2)$, where $\mathcal{V}$ and $\mathcal{E}$ denote the collections of vertices and edges, and a target $\mathbf{t}$ represented as an unsigned distance function $f_\mathbf{t}$,
we define our deformation function $\mathcal{D}(\mathbf{s}; \mathbf{t})$ as follows:
\begin{equation}
    \mathcal{D}(\mathbf{s}; \mathbf{t}) = \Big(\argmin_{\widehat{\mathcal{V}}}  \big\{ \sum_{\hat{v_i}\in\widehat{\mathcal{V}}} f_\mathbf{t}(\hat{v_i}) + \lambda \sum_{(i, j)\in\mathcal{E}} || (\hat{v_i}-\hat{v_j}) - (v_i-v_j) ||^2 \big\}\,, \,\mathcal{E}\Big)
    \label{eq:deform-func}
\end{equation}
\noindent where $v_i$ and $\hat{v_i}$ are the given and optimized positions of $i$-th vertex.
The first term represents the fitting loss that pushes the deformed source shape $\mathcal{D}(\mathbf{s}; \mathbf{t})$ to be close to $\mathbf{t}$, and the second term is the rigidity regularization loss that penalizes for the length changes of each edge in $\mathcal{E}$.
In our implementation, we solve the minimization in Eq.~\ref{eq:deform-func} using Ceres solver~\cite{ceres-solver} by initializing the vertex coordinates with the source mesh and defining the unsigned distance function with $100^3$ voxel grids.
We set $\lambda=1$ in all our experiments, as we found that it well-preserves the CAD model features including sharp edges and corners for most of the 3D models we used.

\begin{figure}[t!]
    \centering
    \includegraphics[width=\linewidth]{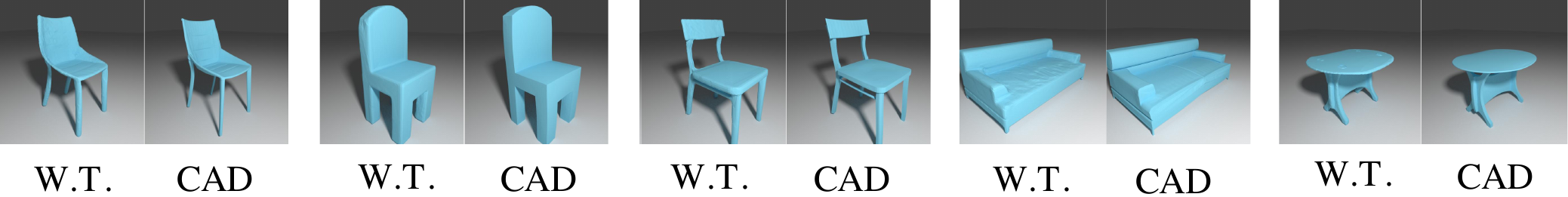}
    \caption{Deformation results with the converted watertight meshes and the raw CAD models.}
    \label{fig:deformation}
\end{figure}

While we convert the CAD models to simplified watertight meshes in Sec.~\ref{sec:implementation_details}~\refpaper{} to efficiently deform and preserve the connectivity across the connected components in the CAD model, the ARAP deformation can also be directly applied to the surface of the CAD model with a simple preprocessing.
We found that remeshing each connected component and linking the components each other with additional edges based on the proximity can also give a very similar result in the deformation with that of using the converted watertight meshes. This way can maintain the original CAD model structure with its accompanied meta information.
Fig.~\ref{fig:deformation} shows the difference between the converted watertight mesh deformation to the direct CAD model deformation, which are almost indistinguishable. All figures of the qualitative evaluation results in our main paper are rendered with the results of the direct CAD model deformation.

\subsection{Image-to-CAD}
\label{sec:exp_image2cad}
\begin{figure}
    \centering
    \includegraphics[width=\linewidth]{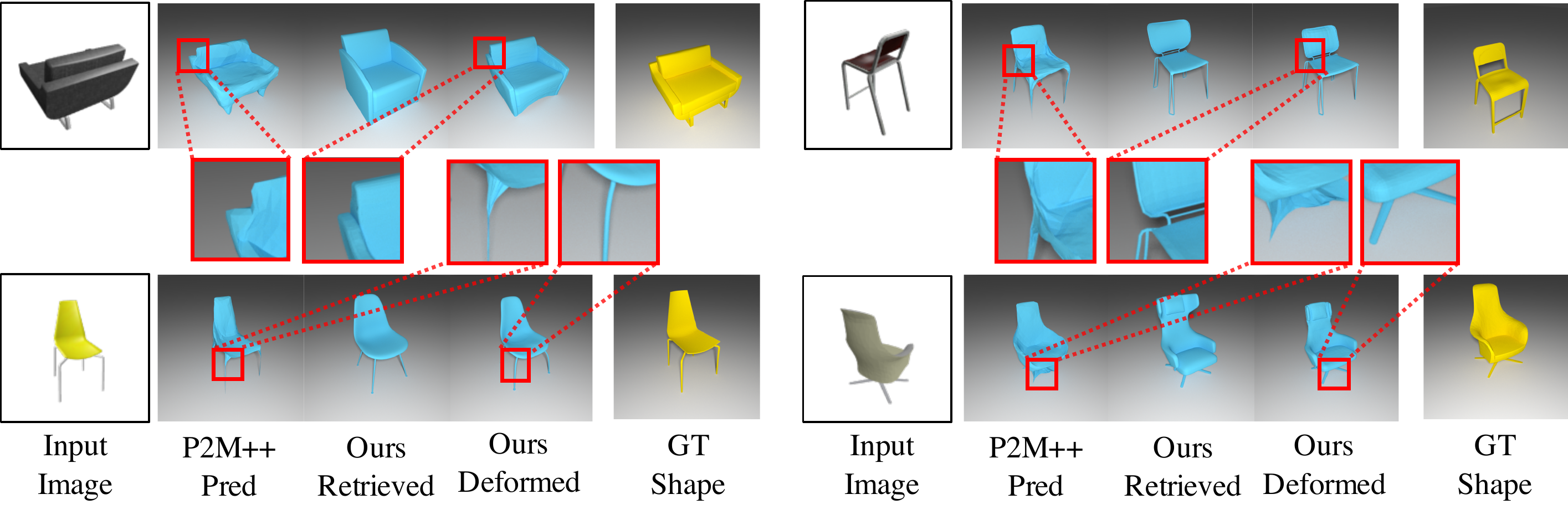}
    \caption{Qualitative results to show the feasibility of our approach for the Image-to-CAD application. We show one of three input viewpoints used by Pixel2Mesh++~\cite{wen2019pixel2mesh++} to produce their coarse mesh. We use this to retrieve a CAD model, which is then deformed to fit the coarse mesh. Rigidity constraints ensure the quality of our output as shown. See Fig.~\ref{fig:image2cad_more} for more results.
    }
    \label{fig:image2cad}
\end{figure}
To show the flexibility of our approach, we now extend it to the application of image-to-CAD generation. Given an image of a 3D model, we first use Pixel2Mesh++~\cite{wen2019pixel2mesh++}, a state-of-the-art image-to-mesh network, to generate an initial coarse mesh. We then use its output to retrieve a CAD model using the proposed \textit{Ours-Reg} trained on ShapeNet~\cite{ShapeNet} and deform it to fit the coarse mesh. Fig.~\ref{fig:image2cad} shows that our approach is able to output models without artifacts produced by direct generation networks \ie in this case Pixel2Mesh++. It is clearly shown that our output has sharper edges and preserves thin structures such as the legs of the chairs in Fig.~\ref{fig:image2cad}. Note that our retrieval solely relies upon the mesh prediction of Pixel2Mesh++~\cite{wen2019pixel2mesh++} and we warp the retrieved model towards the output of this mesh prediction network without the knowledge of the input images.

\subsection{Additional Baselines}

We further compare our method with additional baselines. \emph{CD-Margin} is the case of using the margin loss in Sec.~\ref{sec:method_triplet}~\refpaper{} for training, but employing $d^m$ (Chamfer distance) to define the relationships among the shapes (instead of the fitting gap $e_D^m$ (Eq.~\ref{eq:fitting gap}~\refpaper{}) and using Euclidean distance in the embedding space (fixing $\mathcal{G}(\cdot)$ in Eq.~\ref{eq:egocentric_distance}~\refpaper{} to identity). Given this, we introduce three more baselines:
\begin{itemize}
\item \emph{CD-Reg}: The network is trained with the regression loss in Sec.~\ref{sec:method_regression}~\refpaper{}, but using $d^m$ and identity $\mathcal{G}(\cdot)$ to define shape relationships and embedding distance.
\item \emph{Symm-Margin}: The relationships among the shapes are defined with the fitting gap $e_D^m$ (Eq.~\ref{eq:fitting gap}~\refpaper{}), but still $\mathcal{G}(\cdot)$ in Eq.~\ref{eq:egocentric_distance}~\refpaper{} is fixed to identity.
\item \emph{Symm-Reg}: The same with \emph{Symm-Margin}, but the network is trained with the regression loss in Sec.\ref{sec:method_regression}.
\end{itemize}
The quantitative results are reported in Tab.~\ref{table:add_baseline} and Tab.~\ref{table:add_ranking} (similarly to Tab.~\ref{table:quantitative} and Tab.~\ref{table:ranking}~\refpaper{}). Refer to Sec.~\ref{sec:exp} for the details of the evaluation metrics. The performance is improved when using the fitting gap $e_D^m$ as the relationships instead of Chamfer distance $d^m$, as shown in the results of \emph{Symm-Margin} and \emph{Symm-Reg} (compared with the results of \emph{CD-Margin} and \emph{CD-Reg}). However, still the performance of \emph{Symm-Margin} and \emph{Symm-Reg} is inferior to our case (\emph{Ours-Margin} and \emph{Ours-Reg}) using the egocentric distance field to embed the relationships. Also, note that the regression loss provides better performance only when the egocentric distance field is used in the embedding (\emph{Ours-Margin} vs. \emph{Ours-Reg}) but not for the other cases (\emph{CD-Margin} vs. \emph{CD Reg} and \emph{Symm-Margin} vs. \emph{Symm-Reg}).

\setlength{\tabcolsep}{4pt}
\begin{table}[t!]
\centering
\caption{Additional baseline results that show the fitting gap for the top-1 retrieval of the different object classes in three additional set-ups.The fitting gap $e_\mathcal{D}^m(\mathbf{s}, \mathbf{t})$ multiplied by $1e^{-2}$ are reported.}
\label{table:add_baseline}
{\scriptsize
\begin{tabularx}{\textwidth}{c|>{\centering}m{2cm}|CC|CC|CC|CC}
\toprule
 \multicolumn{2}{c|}{\multirow{2}{*}{Method}} & \multicolumn{2}{c|}{Table} & \multicolumn{2}{c|}{Chair} & \multicolumn{2}{c|}{Sofa} & \multicolumn{2}{c}{Car}  \\
 \multicolumn{2}{c|}{} & {\tiny Top-1} & {\tiny Top-3} & {\tiny Top-1} & {\tiny Top-3} & {\tiny Top-1} & {\tiny Top-3} & {\tiny Top-1} & {\tiny Top-3} \\
\midrule
\multirow{6}{*}{\rotatebox{90}{\makecell{Mean\\$d^m(\mathbf{s},\mathbf{t})$}}}
& CD-Margin &  \textbf{4.875} & \textbf{3.449} & \textbf{4.750} & \textbf{3.518} & \textbf{3.087} & 4.151 & \textbf{2.525} & \textbf{1.905}\\
& CD-Reg  & 9.457 & 5.828 & 9.127 & 5.980 & 7.095 & 4.547 & \underline{2.658} & \underline{1.947} \\
& Symm-Margin  & \underline{5.939} & \underline{3.887} & \underline{5.533} & \underline{3.857} & \underline{4.709} & \textbf{3.301} & 2.958 & 2.137 \\
& Symm-Reg & 6.517 & 4.025 & 9.824 & 6.579 & 7.667 & 4.990 & 2.989 & 2.218 \\
& \textbf{Ours-Margin} &  6.227 & 4.026 &  5.664 & 3.889 & 4.825 & \underline{3.400}  &  2.962 & 2.142 \\
& \textbf{Ours-Reg} &  5.955 & 3.979 & 5.751 & 3.981  & 5.091 & 3.628 & 3.119 & 2.263 \\
\midrule
\multirow{6}{*}{\rotatebox{90}{\makecell{Mean\\$e^m_\mathcal{D}(\mathbf{s},\mathbf{t})$}}}
& CD-Margin & 2.362 & 1.373  & 2.134 & 1.242  &  1.587 & 0.909 &  1.249 & 0.773 \\
& CD-Reg & 5.086 & 2.736 & 4.166 & 2.310 & 3.186 & 1.498 & 1.327 &  0.778 \\
& Symm-Margin  & 2.183  & 1.267  & 1.946 & 1.169 & 1.497 & 0.855 & 1.261 & 0.743  \\
& Symm-Reg & 2.500 & 1.334 & 4.349 & 2.591 & 3.313 & 1.639 & \underline{1.157} & \underline{0.695}  \\
& \textbf{Ours-Margin} & \underline{2.127} & \underline{1.251} & \underline{1.915} & \underline{1.144} & \underline{1.420} & \underline{0.835} & 1.226 & 0.747 \\
& \textbf{Ours-Reg} &  \textbf{1.969} & \textbf{1.129} & \textbf{1.752} & \textbf{1.054} & \textbf{1.338} & \textbf{0.788} & \textbf{1.112} & \textbf{0.681} \\
\bottomrule
\end{tabularx}
}
\end{table}
\setlength{\tabcolsep}{1.4pt}

\setlength{\tabcolsep}{4pt}
\begin{table}[t!]
\centering
\caption{Additional baseline results for ranking evaluations with 150 models per query. The models are randomly selected and sorted by $e^m_\mathcal{D}(\mathbf{s},\mathbf{t})$ (the query is not included). All results are for the top-$1$ retrieval results of each method. The numbers multiplied by $1e^{-2}$ are reported.
\label{table:add_ranking}
}
{\scriptsize
\begin{tabularx}{\textwidth}{l|CCC|CCC|CCC|CCC}
\toprule
\multirow{3}{*}{Method} & \multicolumn{3}{c|}{Table} & \multicolumn{3}{c|}{Chair} & \multicolumn{3}{c|}{Sofa} & \multicolumn{3}{c}{Car}  \\
 & {\tiny Mean} & {\tiny Mean} & {\tiny Mean}
 & {\tiny Mean} & {\tiny Mean} & {\tiny Mean}
 & {\tiny Mean} & {\tiny Mean} & {\tiny Mean}
 & {\tiny Mean} & {\tiny Mean} & {\tiny Mean} \\

 & {\tiny $d^m$} & {\tiny $e^m_\mathcal{D}$} & {\tiny Rank} 
 & {\tiny $d^m$} & {\tiny $e^m_\mathcal{D}$} & {\tiny Rank} 
 & {\tiny $d^m$} & {\tiny $e^m_\mathcal{D}$} & {\tiny Rank} 
 & {\tiny $d^m$} & {\tiny $e^m_\mathcal{D}$} & {\tiny Rank} \\
\midrule
CD-Margin & \textbf{6.77}  & 3.19  & 12.55 & \textbf{6.02} & 2.72  & 13.24 & \textbf{5.07} & 1.93  & 15.76 & \textbf{3.02} & 1.48  & 18.94   \\
CD-Reg & 10.37 & 5.42 & 46.67 & 9.51 & 4.31  & 41.35 & 7.62 & 3.32 & 43.06 & \underline{3.16} & 1.45 & 18.66  \\
Symm-Margin & \underline{8.54}  & 2.96  & 9.70 & \underline{7.09}  & 2.46  & 9.31 & \underline{5.69}  & 1.77  & 11.04 & 3.54 &  1.37 &  14.83 \\
Symm-Reg & 8.72  & 3.15 & 12.56 & 10.37 & 4.61  & 46.54 & 7.62 & 3.32 & 43.06 & \underline{3.16} & 1.45 & 18.66  \\
\textbf{Ours-Margin} & 8.89  & \underline{2.88} & \underline{8.86} & 7.15 & \underline{2.37} & \underline{8.15} &  5.83 & \underline{1.67} & \underline{9.09} &  3.61 & \underline{1.34} & \underline{12.95} \\
\textbf{Ours-Reg} & 8.59  & \textbf{2.71} & \textbf{7.05} & 7.39  & \textbf{2.24} & \textbf{6.32} &  6.23 & \textbf{1.62} & \textbf{7.91} & 3.80  & \textbf{1.24} & \textbf{7.80}  \\
\bottomrule
\end{tabularx}
}
\end{table}%
\setlength{\tabcolsep}{1.4pt}

\subsection{Additional Evaluation Metric - Recall}
We also report recall of the retrieval results. Since the notion of the \emph{correct} match is not defined in our problem, we compute recall@1 by calculating the proportion of the cases when the top-1 retrieval is in the top-5 ranks based on $e_D^m(s, t)$. The results are reported in Tab.~\ref{table:recall}. Ours outperforms the baselines with big margins.


\begin{table}[t!]
\centering
\caption{The percentage of recall@1 for different methods. A correct match is defined as the case when the top-1 retrieval is in the top-5 ranks based on $e_D^m(s, t)$.}
\label{table:recall}
{\scriptsize
\begin{tabularx}{\textwidth}{>{\centering}m{2cm}|C|C|C|C}
\toprule
 Method & Table & Chair & Sofa & Car\\
\midrule
Ranked-CD & 50.50 & 52.52 & 46.91 & 54.26 \\
AE & 54.73 & 54.41 & 49.76 & 43.89 \\
CD-Margin &  51.04 & 50.69 & 44.53 & 39.20 \\
CD-reg & 18.34 & 17.43 & 16.64 & 38.07\\
Symm-Margin & 61.35 & 61.29 & 53.25 & 45.03\\
Symm-Reg & 56.52 & 14.23 & 16.80 & \underline{61.22}\\
\textbf{Ours-Margin} &  \underline{64.26} & \underline{65.55} & \underline{58.32} & 46.73 \\
\textbf{Ours-Reg} & \textbf{70.64} & \textbf{73.97} & \textbf{65.61} & \textbf{67.19} \\
\bottomrule
\end{tabularx}
}
\end{table}

\subsection{Hard Negative Mining in the Margin-Loss-Based Approach}
For our margin-loss-based approach (\emph{Ours-Margin}) described in Sec.~\ref{sec:method_triplet}~\refpaper{}, we also tried hard negative mining~\cite{Schroff:2015} in the network training. For each query, we generate the set of negative samples $\mathbf{N'_t}$ with the 8 hardest negatives in $\mathbf{N_t}$ (the closest to the query by the learned egocentric distance $\delta(\mathbf{t}; \mathbf{s})$) and 5 other randomly selected negatives; the additional random negatives are added to avoid overfitting. For training efficiency, instead of forward-propagating the network for each step to compute the egocentric distance $\delta(\mathbf{t}; \mathbf{s})$, we cache the latent vectors $\mathcal{F}(\cdot)$ and the distance field (PSD) matrices $\mathcal{G}(\cdot)$ for all the models in the database and update them every 10 epochs. The hard negative mining was tested in the fine-tuning, and the network model was first trained in the normal way (with all randomly selected negatives) for 30 epochs. Tab.~\ref{table:hardneg} shows the quantitative results on ShapeNet~\cite{ShapeNet}, indicating that the hard negative mining slightly improves the performance. But, \emph{Ours-Reg} still performs better than \emph{Ours-Margin} in all classes.

\begin{table}[t!]
\centering
\caption{Quantitative comparison of \emph{Ours-Margin} with and without the hard negative mining and \emph{Ours-Reg}, experimented on ShapeNet~\cite{ShapeNet}. The fitting gap $e_\mathcal{D}^m(\mathbf{s}, \mathbf{t})$ multiplied by $1e^{-2}$ are reported. Bold is the smallest.}
\label{table:hardneg}
{\scriptsize
\begin{tabularx}{\textwidth}{>{\centering}m{2cm}|CC|CC|CC|CC}
\toprule
 \multirow{2}{*}{Method}& \multicolumn{2}{c|}{Table} & \multicolumn{2}{c|}{Chair} & \multicolumn{2}{c|}{Sofa} & \multicolumn{2}{c}{Car}\\
 & Top-1 & Top-3 & Top-1 & Top-3 & Top-1 & Top-3 & Top-1 & Top-3 \\
\midrule
 \textbf{Ours-Margin} & 2.127 & 1.251 & 1.915 & 1.144 & 1.420 & 0.835 & 1.226 & 0.747\\
 \textbf{Ours-Margin} w/ hardneg & 2.090  & 1.233 & 1.904 & 1.131 & 1.400 & 0.822 & 1.220 & 0.744 \\
 \textbf{Ours-Reg} &  \textbf{1.969} & \textbf{1.129} & \textbf{1.752} & \textbf{1.054} & \textbf{1.338} & \textbf{0.788} & \textbf{1.112} & \textbf{0.681} \\
\bottomrule
\end{tabularx}
}
\end{table}

\begin{table}[t!]
\centering
\caption{Qualitative comparisons on ShapeNet Table dataset with the varying dimension of the latent space. The fitting gap $e_\mathcal{D}^m(\mathbf{s}, \mathbf{t})$ multiplied by $1e^{-2}$ are reported. Bold is the smallest among the dimensions.}
\label{table:dimensionality}
{\scriptsize
\begin{tabularx}{\textwidth}{>{\centering}m{2cm}|CC|CC|CC|CC}
\toprule
 \multirow{2}{*}{Dimension} & \multicolumn{2}{c|}{AE} & \multicolumn{2}{c|}{CD-Margin} & \multicolumn{2}{c|}{\textbf{Ours-Margin}} & \multicolumn{2}{c}{\textbf{Ours-Reg}}\\
 & Top-1 & Top-3 & Top-1 & Top-3 & Top-1 & Top-3 & Top-1 & Top-3 \\
\midrule
$d=64$  & 2.481 & 1.489 & 2.429 & 1.418 & 2.159 & \textbf{1.229} & 1.997 & 1.153 \\
$d=128$ & \textbf{2.325} & 1.357 & 2.369 & 1.380 & 2.131 & 1.243 & 1.981 & 1.133 \\
$d=256$ & 2.331 & \textbf{1.334} & 2.362 & 1.373 & 2.127 & 1.251 & \textbf{1.969} & \textbf{1.129} \\
$d=512$ & 2.330 & 1.351 & \textbf{2.323} & \textbf{1.370} & \textbf{2.092} & 1.235 & 2.006 & 1.143 \\
\bottomrule
\end{tabularx}
}
\end{table}

\subsection{Analysis of Latent Space Dimension}

As mentioned in Sec.~\ref{sec:implementation_details}~\refpaper{}, we demonstrate the effect of varying the dimension of the latent space, both for baselines and our methods. Tab.~\ref{table:dimensionality} shows the quantitative results on ShapeNet Table dataset when varying the dimension of the latent space from $64$ to $512$. While the higher dimensions mostly offer slightly better performance, the difference is marginal, meaning that even the smallest dimension ($d=64$) has sufficient capacity to encode the asymmetric deformability relationships. Also, regardless of the dimension, our methods consistently outperform the baselines with significant margins.

\clearpage
\subsection{More Qualitative Results}

In the following figures, we show more qualitative comparisons between our method and baseline methods for the experiments of ShapeNet (Sec.~\ref{sec:exp_shapenet}), Scan-to-CAD (Sec.~\ref{sec:exp_scan2cad})~\refpaper{}, and Image-to-CAD (Sec.~\ref{sec:exp_image2cad}).

\begin{figure}[ht!]
    \centering
    \includegraphics[width=\linewidth]{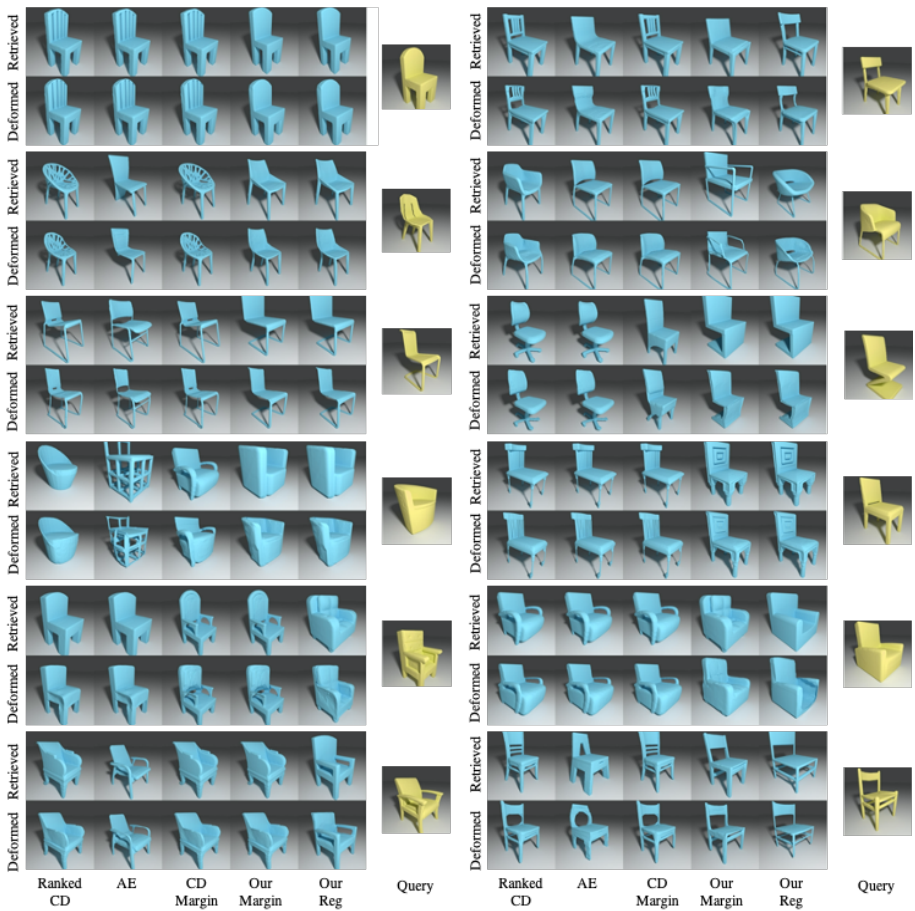}
    \caption{More qualitative results of ShapeNet experiment (chairs). See Sec.~\ref{sec:exp_shapenet}~\refpaper{} for the details.}
    \label{fig:shapenet-more1}
\end{figure}
\clearpage

\begin{figure}[ht!]
    \centering
    \includegraphics[width=\linewidth]{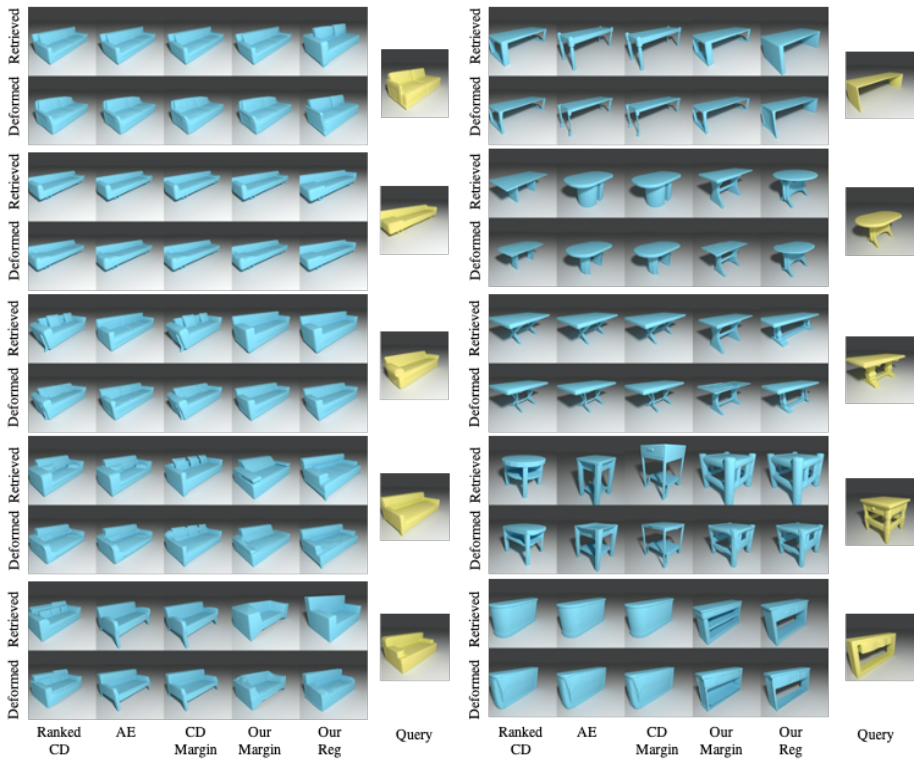}
    \caption{More qualitative results of ShapeNet experiment (sofas and tables). See Sec.~\ref{sec:exp_shapenet}~\refpaper{} for the details.}
    \label{fig:shapenet-more2}
\end{figure}

\begin{figure}[ht!]
    \centering
    \includegraphics[width=\linewidth]{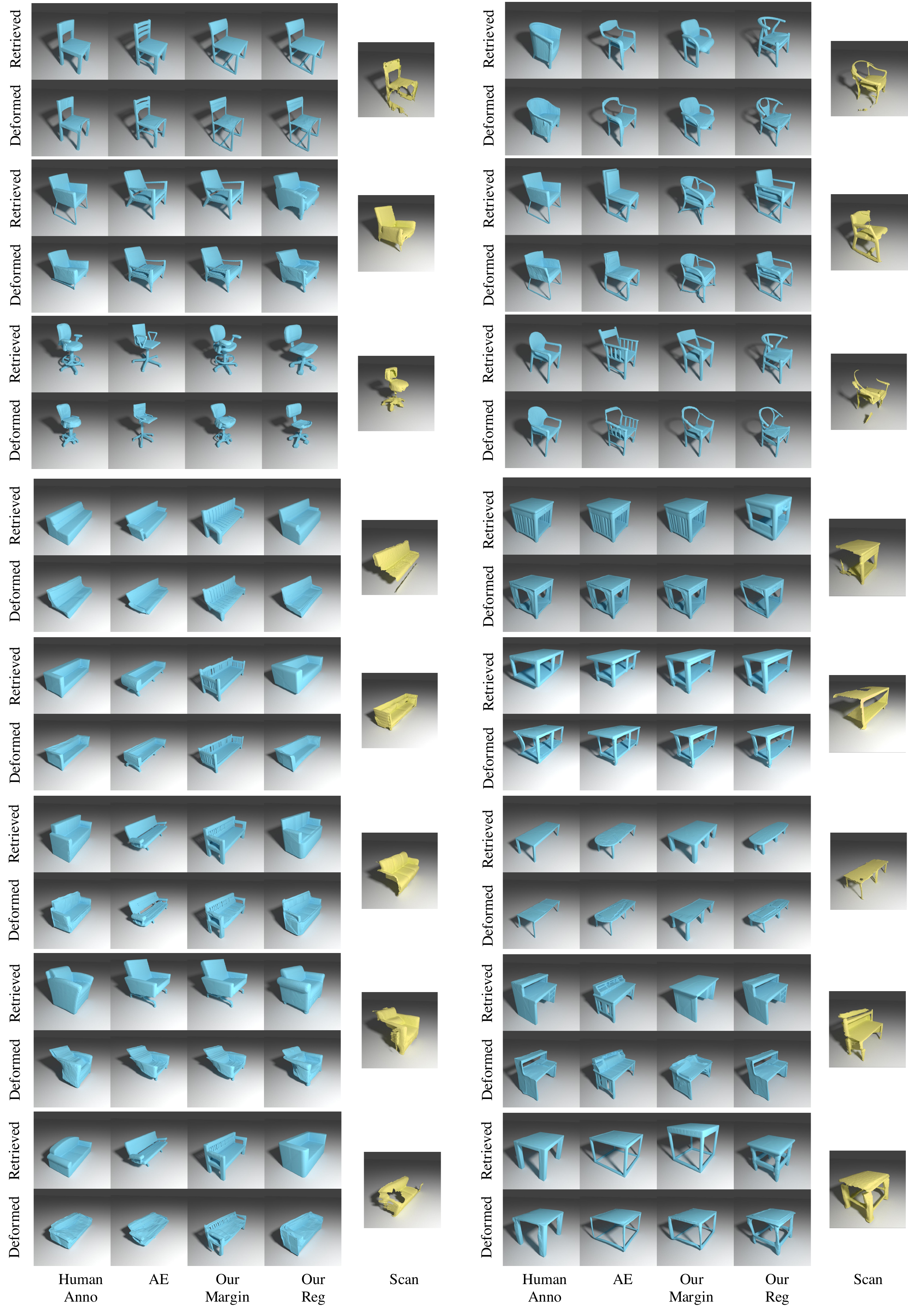}
    \caption{More qualitative results of Scan-to-CAD experiment (chairs, tables, sofas). See Sec.~\ref{sec:exp_scan2cad}~\refpaper{} for the details.}
    \label{fig:scan2cad-more1}
\end{figure}
\clearpage

\begin{figure}[ht!]
    \centering
    \includegraphics[width=\linewidth]{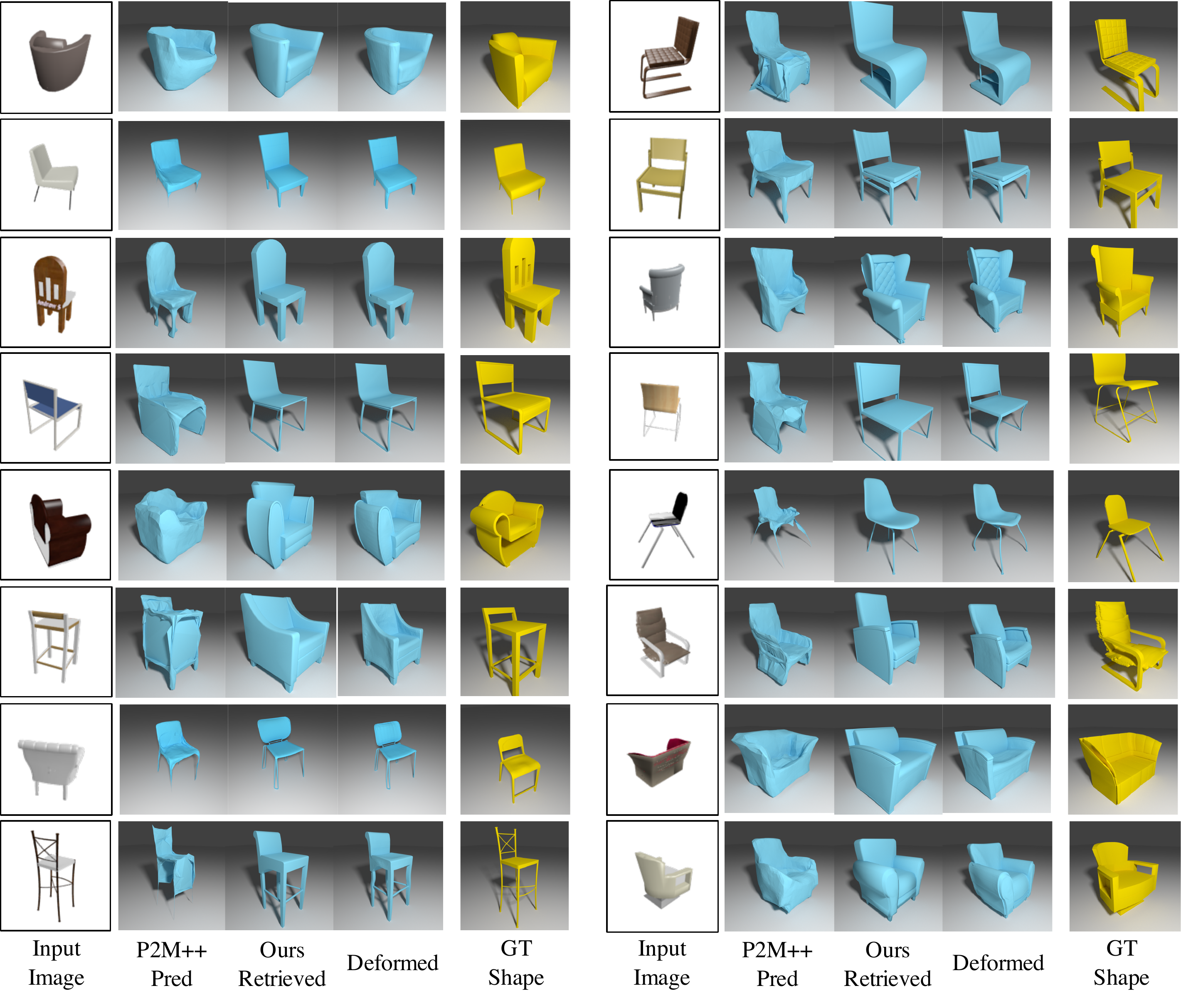}
    \caption{More qualitative results of Image-to-CAD experiment. See Sec.~\ref{sec:exp_image2cad} for the details.}
    \label{fig:image2cad_more}
\end{figure}

\end{document}